\documentclass[lettersize,journal]{IEEEtran}
\usepackage{caption}
\usepackage{multirow}
\usepackage{booktabs}
\usepackage{tabularx}
\usepackage{diagbox}
\usepackage{amsmath,amsfonts}
\usepackage{algorithmic}
\usepackage{algorithm}
\usepackage{array}
\usepackage{color}
\usepackage{mathrsfs}
\usepackage{times}
\usepackage{balance}
\usepackage{graphicx}
\usepackage{enumerate}
\usepackage{amsmath,amsfonts}\usepackage[caption=false,font=normalsize,labelfont=sf,textfont=sf]{subfig}
\usepackage{textcomp}
\usepackage{stfloats}
\usepackage{url}
\usepackage{verbatim}
\usepackage{graphicx}
\usepackage{cite}
\usepackage[hidelinks]{hyperref}
\definecolor{mycustompurple}{RGB}{154, 36, 79} 
\usepackage[utf8]{inputenc}
\hypersetup{
    colorlinks=true,            
    linkcolor=red,              
    citecolor=blue,             
    filecolor=mycustompurple,   
    urlcolor=magenta            
}

\begin{document}

\title{MsMemoryGAN: A Multi-scale Memory GAN for Palm-vein Adversarial Purification}
\author{Huafeng Qin, Changqing Gong,
        Yantao Li, Mounim~A.~El-Yacoubi,  Xinbo Gao, Qun song, and Jun Wang     

}

\author{Huafeng Qin, Yuming Fu, Huiyan Zhang, Mounim~A.~El-Yacoubi,  Xinbo Gao, ~\IEEEmembership{Fellow, IEEE}, Qun Song,and Jun Wang    
\thanks{H. Qin, Y. Fu,  H.Zhang, and Qun song are with Chongqing Technology and Business University, Chongqing 400067, China (e-mail: qinhuafengfeng@163.com, fym291715@163.com).} 
\thanks{M. A. El-Yacoubi is with SAMOVAR, Institut Polytechnique de Paris, 91120 Palaiseau, France (e-mail: mounim.el\_yacoubi@telecom-sudparis.eu).}
\thanks{ X. Gao is with the Chongqing University of Posts and Telecommunications, Chongqing 400065, China (e-mail: gaoxb@cqupt.edu.cn).}
\thanks{ J. Wang is with the China University of Mining and Technology, Jiangsu 221116, China (e-mail:WJ999LX@163.com).}
\thanks{Manuscript received September XX, 2023; revised XXXX XX, 201X. 
(Corresponding author: Qun Song and Huiyan Zhang )} 
\thanks{Manuscript received April 19, 2021; revised August 16, 2021.}}

\markboth{Journal of \LaTeX\ Class Files,~Vol.~14, No.~8, February~2024}%
{Shell \MakeLowercase{\textit{et al.}}: A Sample Article Using IEEEtran.cls for IEEE Journals}


\maketitle

\begin{abstract}
Deep neural networks have recently achieved promising performance in the vein recognition task and have shown an increasing application trend, however, they are prone to adversarial perturbation attacks by adding imperceptible perturbations to the input, resulting in making incorrect recognition. To address this issue, we propose a novel defense model named MsMemoryGAN, which aims to filter the perturbations from adversarial samples before recognition.  
First, we design a multi-scale autoencoder to achieve high-quality reconstruction and two memory modules to learn the detailed patterns of normal samples at different scales. 
Second, we investigate a learnable metric in the memory module to retrieve the most relevant memory items to reconstruct the input image.  
Finally, the perceptional loss is combined with the pixel loss to further enhance the quality of the reconstructed image. 
During the training phase, the MsMemoryGAN learns to reconstruct the input by merely using fewer prototypical elements of the normal patterns recorded in the memory. At the testing stage, given an adversarial sample, the MsMemoryGAN retrieves its most relevant normal patterns in memory for the reconstruction. Perturbations in the adversarial sample are usually not reconstructed well, resulting in purifying the input from adversarial perturbations. We have conducted extensive experiments on two public vein datasets under different adversarial attack methods to evaluate the performance of the proposed approach. The experimental results show that our approach removes a wide variety of adversarial perturbations, allowing vein classifiers to achieve the highest recognition accuracy.
\end{abstract}


\section{Introduction}

With the rapid development and wide application of Internet technology, information security has received tremendous attention in the past years. Currently, there are various identification/verification technologies such as passwords, cards, fingerprints, and face recognition. Traditional Authentication methods such as passwords, cards, and keys fail to meet the users' requirements in terms of high security and convenience. By leveraging humans' physiological (e.g., face, fingerprint) or behavioral (e.g., voice, gait) characteristics for identification/verification, biometric technologies have become the solution of choice for authentication, with several systems commercially deployed and numerous biometric modalities extensively investigated over the last years.

Biometrics traits can be broadly categorized into two classes \cite{li2023transformer}\cite{7890487}: (1) External modalities, such as face \cite{139758,liu2020hallucinating}, fingerprint \cite{587996,liu2021fingerprint}, and iris \cite{1262028}; (2) Internal modalities, i.e. finger vein \cite{7890487}, hand vein \cite{2009Personal,huang2014hand} and palm vein \cite{9354642}.  The external traits are located on the body surface, making the related authentication techniques susceptible to attacks \cite{2021Attacks}. Facial and fingerprint features, for instance, can be copied without users' consent and the fake version may be employed to fool the face recognition system \cite{2015Deep}\cite{2021Research}. The usage of external traits, as a result, raises serious privacy and security issues. Vein biometrics, by contrast, has the following advantages \cite{2019Finger}\cite{2019A}: (1) High security and privacy: Vein blood vessels are hidden inside the body, making them very hard to copy or steal without user awareness. Also, it is difficult to forge a fake vein vessel for spoofing attacks; (2) Liveliness identification: Vein patterns are captured by NIR light with a wavelength of about 850nm. When the NIR light penetrates the skin, the hemoglobin in the blood absorbs more NIR light than surrounding tissues, making veins appear as darker lines or shadows in the acquired images. As hemoglobin only exists in lively bodies, vein biometrics is a natural liveliness identification technology. These reasons explain the dramatic increase of research works on vein recognition in recent years.

\subsection{Motivation}
\begin{figure}[h!]
    \centering
    \includegraphics[width=0.98\linewidth]{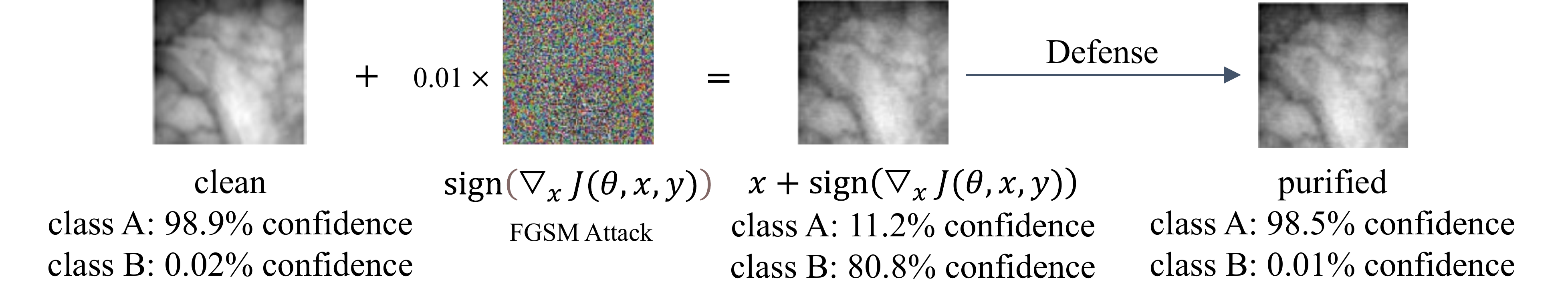}
    \caption{Adversarial attack results. The perturbation is generated by FSGM attack \cite{goodfellow2014explaining} with Vit \cite{dosovitskiy2010image} classifier. Class A represents the correct class and class B represents the incorrect class. The confidence score of the original image belongs to class A is 98.9\%, after adding perturbations to the original image, the resulting adversarial image is classified to class A with a confidence score of 11.2\% while it is misclassified to class B with 80.8\% probability. After feeding it into our model for purification, the confidence of the resulting purifier images belonging to class A is increased to 98.5\%}.
    \label{attacksample}
\end{figure}

Various approaches \cite{61ebd9875244ab9dcb281f13} have been proposed for vein recognition recently. They can be broadly split into two categories. (1) Traditional vein recognition approaches, i.e, handcrafted-based approaches \cite{2007Extraction} \cite{han2012palm} 
and traditional machine learning-based approaches  \cite{li2021joint} \cite{yang2021finger}. 
Typical handcrafted approaches employ handcrafted descriptors such as curvature \cite{2007Extraction}, Gabor \cite{han2012palm} and LBP \cite{liu2017customized}, to extract the vein patterns. 
Differently, some works introduced sparse coding \cite{li2021joint}, low-rank matrix \cite{yang2021finger} and {SVM} \cite{veluchamy2017system}, PCA \cite{kamaruddin2019new}, to automatically learn the vein patterns, avoiding the need of first explicitly extracting some image processing-based features that might discard relevant information about the recognition. (2) Deep learning-based vein recognition approaches. Deep learning (DL) has been proven to be a very powerful tool \cite{2021A} \cite{2012ImageNet} 
and shown super feature representation capacity in various computer vision tasks such as image recognition \cite{2012ImageNet}, data augmentation \cite{iclr2024adautomix}, object tracking \cite{2016Fully}, image segmentation \cite{2021Image}, and so on. It is not surprising then that deep neural networks have been widely applied to vein recognition \cite{yang2019fv}\cite{yang2020fvras}\cite{pan2020multi}\cite{lu2021novel}, with promising performance. Although recent works 
\cite{yang2023downsampling}\cite{yang2023multi} show the domination of DL models in vein recognition tasks, \cite{2020Adversarial} \cite{goodfellow2014explaining} \cite{madry2018towards} have demonstrated that the DL models are vulnerable to adversarial attacks, which, by adding human-imperceptible perturbations to the original inputs, can mislead classifiers into mis-classifying the perturbed inputs. As shown in Fig. \ref{attacksample}, the DL-based classifier Vit  \cite{dosovitskiy2010image} is trained based on normal samples and the trained model recognizes a clean sample input with a 98.9\% confidence level of belonging to its actual class A. When attacking it through an adversarial generator FGSM \cite{goodfellow2014explaining} by adding adversarial perturbations, however, the resulting adversarial sample is misclassified into class B with a high confidence level (80.8\%), even though the perturbation noise is nonperceivable to a human observer. It is possible, therefore, that the attacker intentionally modified data to access genuine users' IDs, resulting in a significant degradation of vein recognition systems’ security.  

To overcome this problem, some researchers have developed various approaches \cite{nie2022diffusion}\cite{samangouei2018defense} to defend deep learning-based recognition models against adversarial attacks. For the same purpose, Li et. al \cite{li2023transformer} made a first attempt to propose a transformer-based defense GAN, named VeinGuard, for vein adversarial perturbation purification. In \cite{li2023transformer}, a purifier consisting of a trainable residual network and a pre-trained generator was trained by minimizing the reconstruction error, to remove a wide variety of adversarial perturbations. The purifier can be seen as an AutoEncoder(AE) \cite{Bengio2013Representation} as the trainable residual network and the pre-trained generator can be treated as an encoder and a decoder, respectively. AE \cite{Bengio2013Representation} is a powerful model for learning high-dimensional data representations in an unsupervised setting. The encoding acts essentially as an information bottleneck which encourages the network to learn the representative patterns of high-dimensional data. In the adversarial defense context, the AE, usually trained on normal samples by minimizing the reconstruction error, uses the trained model as a purifier to remove perturbations. We expect, generally, that the normal samples can be effectively reconstructed. For adversarial samples, the vein patterns only, and not the perturbations, are usually well reconstructed. In other words, the reconstruction error will be small for normal samples and large for adversarial samples. Recent work \cite{2020Memorizing}, however, has implied that sometimes the AEs may have good “generalization”  so that they are also capable of well reconstructing the abnormal samples. The adversarial samples, therefore, may also be well reconstructed by AEs, failing thereby to purify the adversarial perturbations. To mitigate the drawbacks of AEs, a deep autoencoder with a memory module (MemAE) \cite{2020Memorizing} was proposed for anomaly detection. Given an input, MemAE does not directly forward its latent encoding representation into the decoder for reconstruction. Rather, it combines its most relevant items in the memory as its encoding, which is input to the decoder for reconstruction. All the normal samples are used for training, enabling thereby the memory to record the prototypical normal patterns in the normal training data. In the test phase, an abnormal sample is also reconstructed by using fewer patterns from the normal samples. The reconstructed sample, however, tends to be close to the normal data as the adversarial image is reconstructed through a linear combination of normal images. In other words, there is a large reconstruction error for abnormal samples, which can be used as a criterion to detect the anomalies. Memory-based approaches have recently been applied to anomaly detection \cite{2020Memorizing} \cite{chen2020mama} 
and shown state-of-the-art performance. 

\subsection{Our work}
Inspired by Memory-based approaches' success in anomaly detection, we propose a multi-scale memory-augmented GAN (MsMemoryGAN) to defend the vein recognition model against adversarial perturbation attacks. 
 {First, we propose a multi-scale memory-augmented autoencoder to reconstruct the input images with high quality by modeling multi-scale hierarchical features. Then we combine the autoencoder with a discriminator to form MsMemoryGAN, where the goal of the autoencoder is to reconstruct the input samples with high quality so that the discriminator is unable to distinguish between reconstructed and real samples, and the goal of the discriminator is to correctly judge reconstructed and real samples.}
Secondly, a learnable metric is investigated to compute the {correlation} between the features extracted by the encoder and prototypical normal patterns in the memory module. For reconstruction, based on the resulting correlation, the most relevant patterns in the memory to the input are searched as the input representation. 
{Finally, we combine the perceptional loss and pixel loss for model training to reconstruct high-quality images. This reconstruction image is taken as the purified version of the input and is fed to the deep learning-based classifier for vein recognition. }
The main contributions of our work can be summarized as follows:

   (1)We introduce the memory concept in the vein biometrics context to defend deep learning-based vein 
    {recognition models} against adversarial palm-vein image attacks, and propose a Multi-scale {Memory-augmented Autoencoder} 
    to filter adversarial perturbations.
    
   (2) We further design a GAN framework (MsMemoryGAN) which consists of our proposed multi-scale memory-augmented autoencoder and a patch-based discriminator. Furthermore, adversarial loss and perceptual loss are incorporated into the training stage to enhance the quality of the reconstructed images.
    
   (3) {We design a learnable metric to learn the correlation between the features extracted by the encoder and the normal patterns in the memory module.}
   
   (4) We have conducted extensive experiments on two public palm vein datasets to assess the proposed MsMemoryGAN. The results show that MsMemoryGAN can filter the perturbations and allow the vein recognition model to achieve state-of-the-art results under different adversarial attacks.
\section{Related Works}

\subsection{Traditional Vein Recognition Algorithms}
Traditional vein recognition algorithms can be categorized into hand-designed methods and traditional machine learning (ML) algorithms. The former directly extracts handcrafted features from images for classification. Miura et al. \cite{2007Extraction}, for example, computed the local maximum curvature of the cross-sectional contour of a vein image. 
In 2021, Li et al. \cite{9772684}  proposed a novel compact multi-representation feature to describe the informative vein features in local patches for finger-vein feature representation. For traditional ML algorithms, Veluchamy et al. \cite{veluchamy2017system} proposed a $k$-SVM technique for finger-vein identification, while Nurul et al. \cite{kamaruddin2019new} proposed a new filter CCA which takes into account the basic features of the image. Similarly, Lu et al. \cite{yang2021finger} proposed a low-rank representation to extract as much noise-free discriminative information as possible from finger vein images. 

\subsection{Deep learning-based Vein Recognition Algorithms}
Deep learning (DL) models have shown powerful feature representation capacity and have been successfully applied in various fields such as image recognition \cite{2012ImageNet}, data augmentation \cite{iclr2024adautomix}, object tracking \cite{2016Fully}, image segmentation \cite{2021Image} and so on. DL has also become the dominant approach for vein recognition. Das et al. \cite{das2018convolutional}, for instance, proposed a CNN to achieve stable and high finger vein recognition, while Yang et al. \cite{yang2019fv}, proposed FV-GAN, a new method for finger vein extraction and verification using GANs. Yang et al. \cite{yang2020fvras} then proposed FVRAS-Net, a lightweight CNN integrating both the recognition task and the anti-spoofing task. Qin et al. \cite{9354642} proposed a single-sample single-person (SSPP) palm vein recognition method, while Pan et al. \cite{pan2020multi} designed a multiscale deep representation aggregation model for vein recognition. In 2021, Lu et al. \cite{lu2021novel} proposed a visual transformer (Vit) for finger vein recognition, while, in the same year, Wang et al. \cite{wang2021finger} a multi-receiver field bilinear convolutional NN for the same task, and Hou et al. \cite{hou2021arcvein} proposed a new loss function, called the inverse cosine center loss, to improve the discriminative power of CNNs. In 2022, Shaheed et al. \cite{shaheed2022ds} proposed a pre-trained model based on depth-separable convolutional layers for vein recognition. 
In 2023, Zhang et al. \cite{zhang2023convolutional} proposed a lightweight model and a feature integration model to save learning time while achieving high accuracy. 

\subsection{Defense Methods}
To improve the security of vein recognition systems, some researchers have investigated various defense approaches such as template protection \cite{2015Alignment} \cite{shao2022template} and fake vein detection \cite{qiu2017finger}. Qiu et al., for instance, \cite{qiu2017finger} proposed a novel finger vein attack detection scheme, combining total variation regularization and local binary pattern (LBP) descriptors to improve discrimination and generalization. Lu et al. \cite{2015Alignment} blended 2DPalmHash Code (2DPHC), a cancelable biometric scheme, and Fuzzy Vault primitive to jointly protect palmprint templates, while Shao et al. \cite{shao2022template} proposed a chaotic map-based finger vein template protection method. 

While DL models have been widely applied recently for vein recognition, existing studies have shown that classification models relying solely on DL are vulnerable to adversarial sample attacks \cite{2020Adversarial}. An attacker can attack a DL model through white-box attacks such as FSGM \cite{goodfellow2014explaining} and PGD \cite{madry2018towards} or black-box attacks such as HSJA \cite{chen2020hopskipjumpattack} and SPSA \cite{uesato2018adversarial}. The aforementioned defense methods, however, have not yet been defended against such attacks.  To overcome this problem,  numerous adversarial defense methods have been proposed. Meng et al., for instance, proposed MagNet \cite{meng2017magnet}, a framework for defense against adversarial examples of NN classifiers. Song et al. devised a method named PixelDefend \cite{song2017pixeldefend} to purify images by restoring perturbed images, while Samangouei et al. proposed DefenseGAN \cite{samangouei2018defense}, a defense framework that utilizes a generative model to protect a deep NN from attacks. Inspired by the recent success of Diffusion models in the field of image generation, \cite{nie2022diffusion} introduced a diffusion model for defense by adding Gaussian noise to an attacked image, and then performing a pre-trained inverse diffusion process to recover clean samples. To improve the security of the vein recognition system, Li et al. proposed VeinGuard \cite{li2023transformer}, a model to defend vein classifiers against adversarial vein image attacks.


\begin{figure}[h!]
    \centering
    \includegraphics[width=1\linewidth]{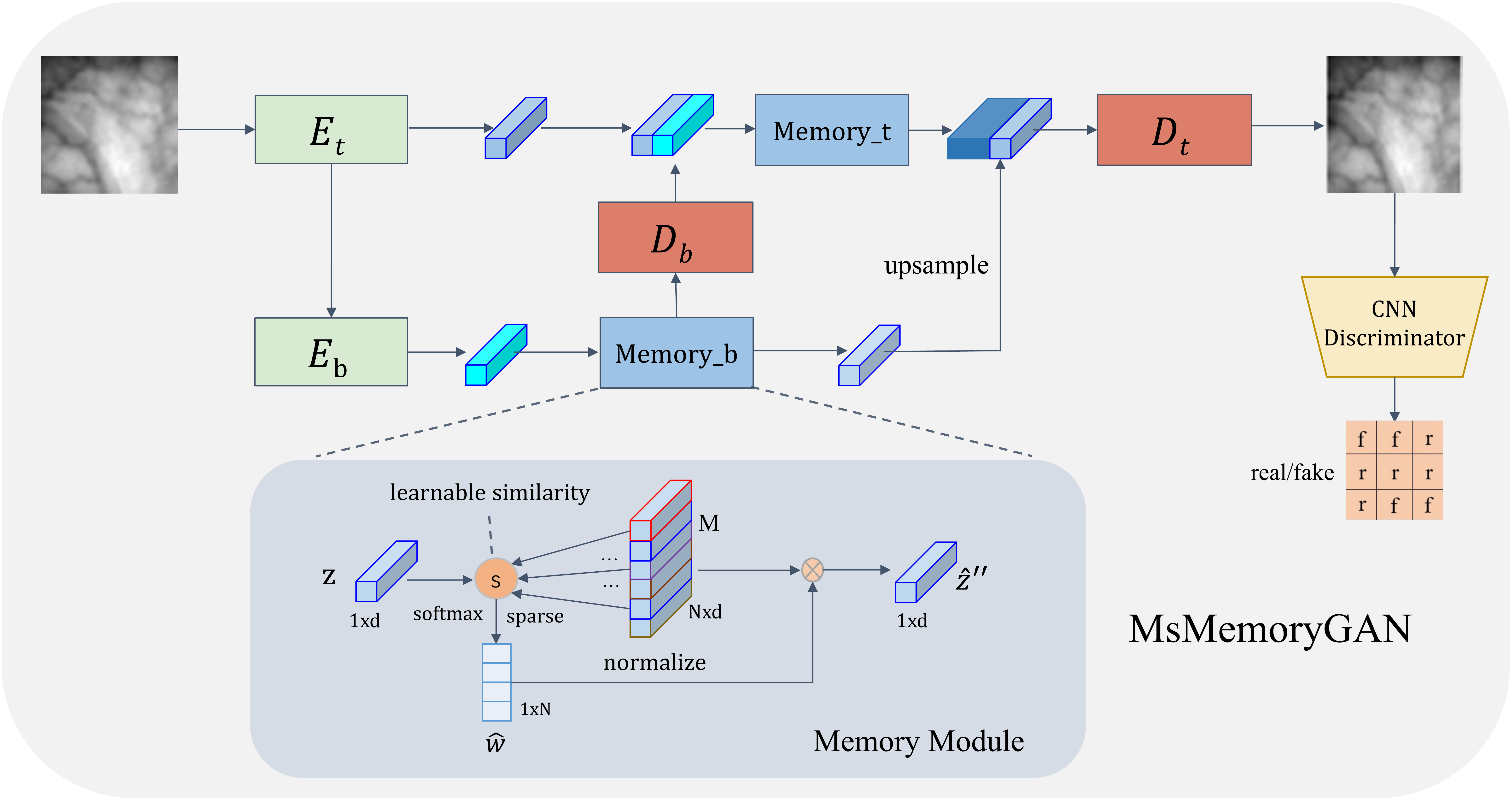}
    \caption{The architecture of the proposed MsMemoryGAN}
    \label{OverallFramework}
\end{figure}

\section{Methods}
Memory based autoencoders  \cite{2020Memorizing} \cite{chen2020mama} \cite{2020Learning} \cite{2021Memory} have been extensively used for anomaly detection.  The original MemAE consists of an encoder that encodes the input into a latent hidden vector, a memory module that retrieves, from the memory, the most relevant entries of the encoder's output to the entries by an attention-based addressing algorithm, and a decoder network to reconstruct the original image. These models suffer from the following problems: (1) Models are trained with a pixel loss such as the $L2$ paradigm to calculate the reconstruction error between pixels. Per-pixel loss, however, does not capture perceptual differences between output and ground-truth images \cite{johnson2016perceptual}, which leads to a blurred reconstructed image; (2) They employ handcrafted metrics such as $cosine$ similarity to retrieve the memory items that are most similar to latent vector $z$ to obtain its representation. These handcrafted metrics may not effectively compute the difference between two latent vectors; (3) As existing approaches process the image at a single scale, they fail to capture local information, such as texture, and global information, such as object shape and geometry, resulting in low-quality images. To defend deep NN-based vein classifiers against adversarial attacks, we propose MsMemoryGAN, a new multi-scale Memory-augmented autoencoder-based defense model to reconstruct clear original images. First, we investigate a multi-scale hierarchical autoencoder to learn the global and local information. Specifically, a top encoder aims to model global information, while a bottom encoder, conditioned on the top latent code is responsible for representing local details. For image reconstruction, the decoder takes both latent codes as input. Second, we propose an improved memory module to retrieve the most relevant items in the memory via a learnable metric. Finally, we introduce a perceptual loss and an adversarial loss, instead of the $ L2$ loss, for reconstruction. The encoder and decoder are trained to minimize the reconstruction error, while the memory contents are simultaneously encouraged to record the prototypical elements of the encoded normal data, to obtain a low average reconstruction error. During the test, the model merely uses a limited number of the normal patterns recorded in the memory to perform the reconstruction. As a result, we usually get small reconstruction errors for normal samples and large errors for adversarial samples. In other words, our approach is capable of purifying the perturbations from adversarial samples.

\subsection{Multi-Scale Memory AutoEncoder}
\label{msmemae}
\begin{figure}[h!]
    \centering
    \includegraphics[width=0.9\linewidth]{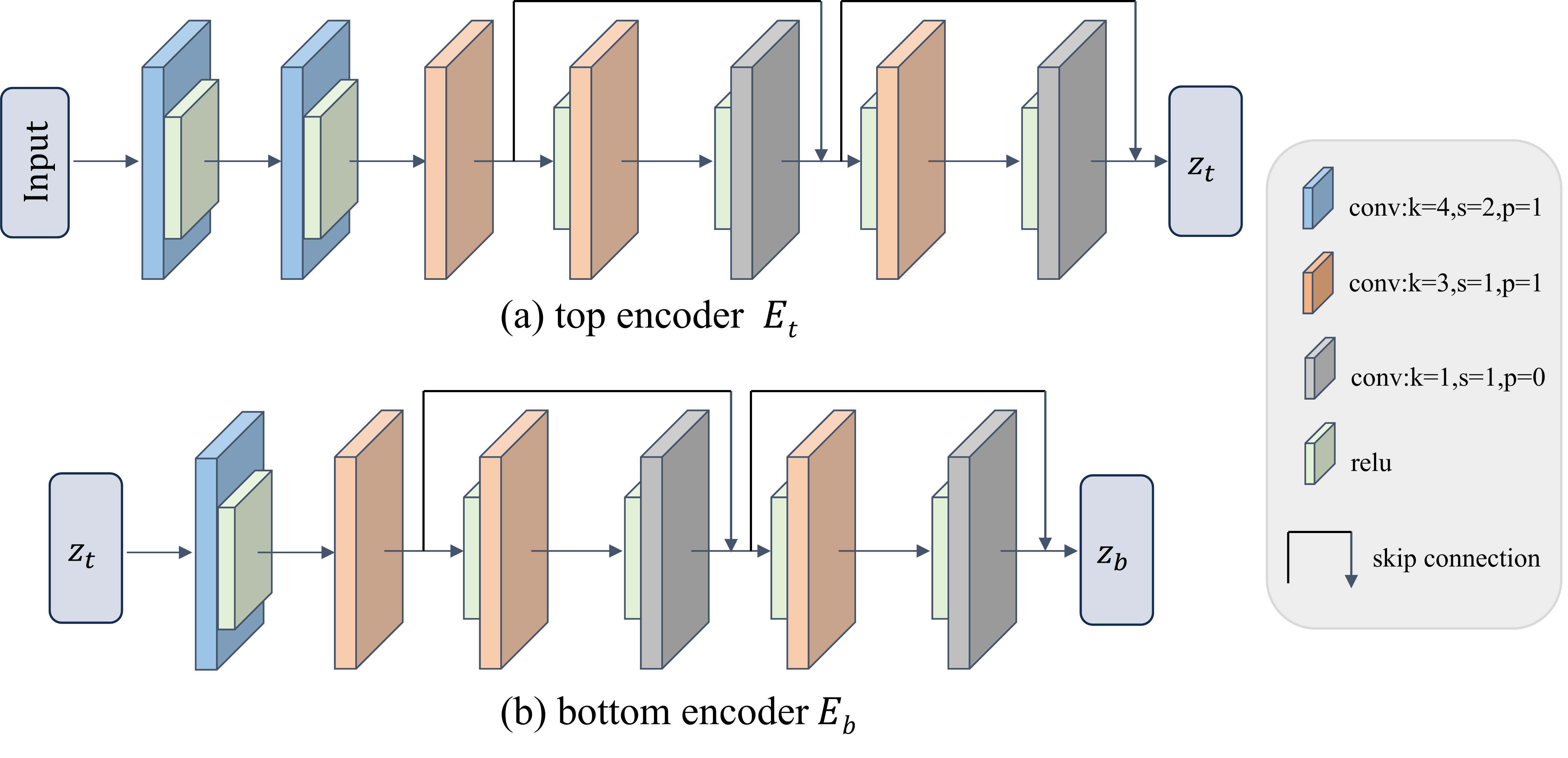}
    \caption{The architecture of top encoder (a) and bottom encoder (b). The former consists of seven convolutional layers while the latter includes six convolutional layers. Some layers in both encoders are stacked with residual connections. The top encoder transforms and downsamples the image by a factor of 4, while the bottom encoder transforms and downsamples the image by a factor of 2.}
    \label{encoder}
\end{figure}

\begin{figure}[h!]
    \centering
    \includegraphics[width=0.9\linewidth]{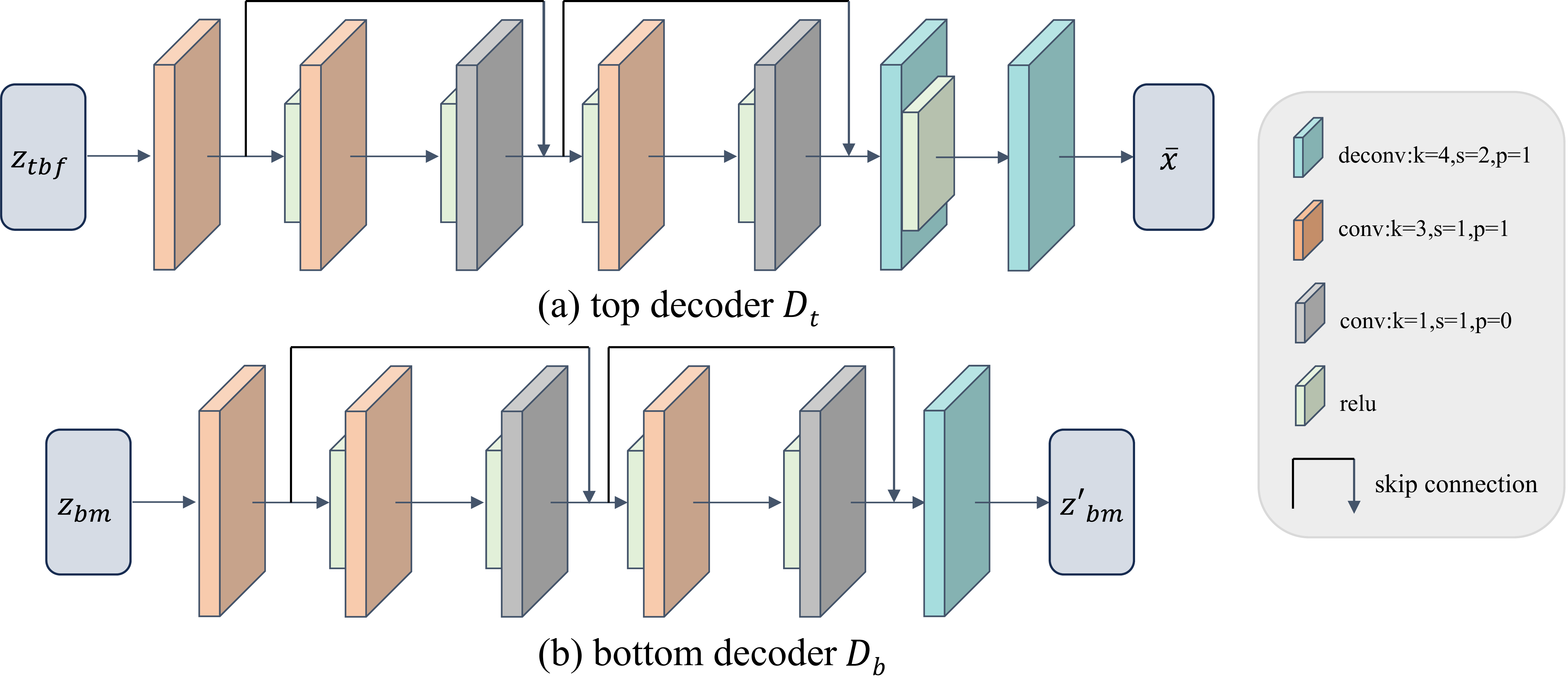}
    \caption{The architecture of top decoder (a) and bottom decoder (b). The former consists of seven convolutional layers, while the latter includes six convolutional layers. Note that some convolutional layers are stacked with residual connections. The top decoder transforms and upsamples the input vector by a factor of 4, while the bottom decoder transforms and upsamples the input by a factor of 2.}
    \label{decoder}
\end{figure}

To purify adversarial perturbations, we propose a multi-scale Memory AutoEncoder, as shown in Fig. \ref{OverallFramework}, which consists of two memory modules, two encoders modules, and two decoders modules. The two encoders encode the input image at two scales to obtain the local details and global information. The memory modules aim to retrieve the most relevant patterns in the memory for input, to obtain its latent representation for reconstruction. The two decoders are responsible for reconstructing the image from the resulting latent representation. Let $x\in\mathbb{R}^{H\times W\times C}$ be a normal vein sample, where $W$ and $H$ are the input image width and height, and $C$ is the number of channels.  As shown in Fig. \ref{encoder}(a),  the top encoder $E_t$ includes seven convolutional layers, followed by a Relu activation function. The first two convolutional layers perform convolutional operations with a stride of 2, which is equivalent to transform and downsample the input image by a factor of 4.  The bottom encoder (Fig.3(b)) $E_b$ consists of six convolutional layers. As the first layer includes a convolutional operator with a stride of 2, the input image is subject to downsampling with a factor of 2. Taking $x$ as the input, the top encoder outputs the local detail feature representation $z_{t}\in\mathbb{R}^{\frac{H}{4}\times\frac{W}{4}\times C_{t}}$, which is further fed to the bottom encoder to learn global feature representation $z_b\in\mathbb{R}^{\frac H8\times\frac W8\times C_b}$. The feature representations are computed by Eq. (\ref{1}) and Eq. (\ref{2}):
\begin{equation}z_t=E_t(x), \label{1}\end{equation}
\begin{equation}z_b=E_b(z_t),\label{2}\end{equation}
where $E_t$ and $E_b$ denote the top encoder and bottom encoder, as shown in Fig. \ref{encoder}. 
The bottom latent code $z_b$ is mapped to a new feature representation $z_{bm}$ through the bottom memory module $Memory_b$, which is input to the bottom decoder $D_b$ for reconstruction. As shown in Fig. \ref{decoder}(b), the bottom decoder $D_b$ includes a convolutional layer, two residual modules, and a deconvolutional layer with a stride of 2. The bottom decoder $D_b$ transforms and upsamples the input map to get purifying feature map $z'_{bm} \in \mathbb{R}^{\frac{H}{4} \times \frac{W}{4} \times C_{t}}$, which is calculated by  Eq. (\ref{3}):

\begin{equation}z'_{bm}=D_b(Memory_b(z_b)), \label{3}\end{equation}
where $memory_b$ is the memory module to purify the adversarial perturbations from the attacked sample, to be detailed in the following section. The resulting feature representation $z'_{bm}$ is combined with latent code $z_t$ in Eq. (\ref{1}) to obtain the $z_{tf}\in\mathbb{R}^{\frac{H}{4}\times\frac{W}{4}\times2C_{t}}$ by Eq. (\ref{4}):
\begin{equation}z_{tf}=\mathrm{Concat}[z_t,z'_{bm}], \label{4}\end{equation}

The resulting top feature $z_{tf}$ is then input into the top memory module $Memory_b$ to remove
the perturbations of samples and the resulting $z_{tm}$ is obtained by Eq. (\ref{5}):
\begin{equation}z_{tm}=Memory_t(z_{tf}), \label{5}\end{equation}

Finally, the purified features $z_{tm}$ and $z_{bm}$ are fused to get the multi-scale hierarchical features $z_{tbf}\in\mathbb{R}^{\frac{H}{4}\times\frac{W}{4}\times{3C}_ t}$, computed by Eq. (\ref{6}):
\begin{equation}z_{tbf}=\mathrm[{Concat}[z_{tm},deconv(z_{bm})]], \label{6}\end{equation}
where $deconv$ denotes the inverse convolution operation to upsample $z_{bm}$. The resulting feature representation $z_{tbf}$ is fed to the top decoder $D_t$ for image reconstruction. The reconstructed image  $\bar{x}\in R^{H\times W\times C}$ is computed by Eq. (\ref{7}):
\begin{equation}\bar{x}=D_t(z_{tbf}), \label{7}\end{equation}

\subsection{ Memory module }
\begin{figure}[h!]
    \centering
    \includegraphics[width=0.8\linewidth]{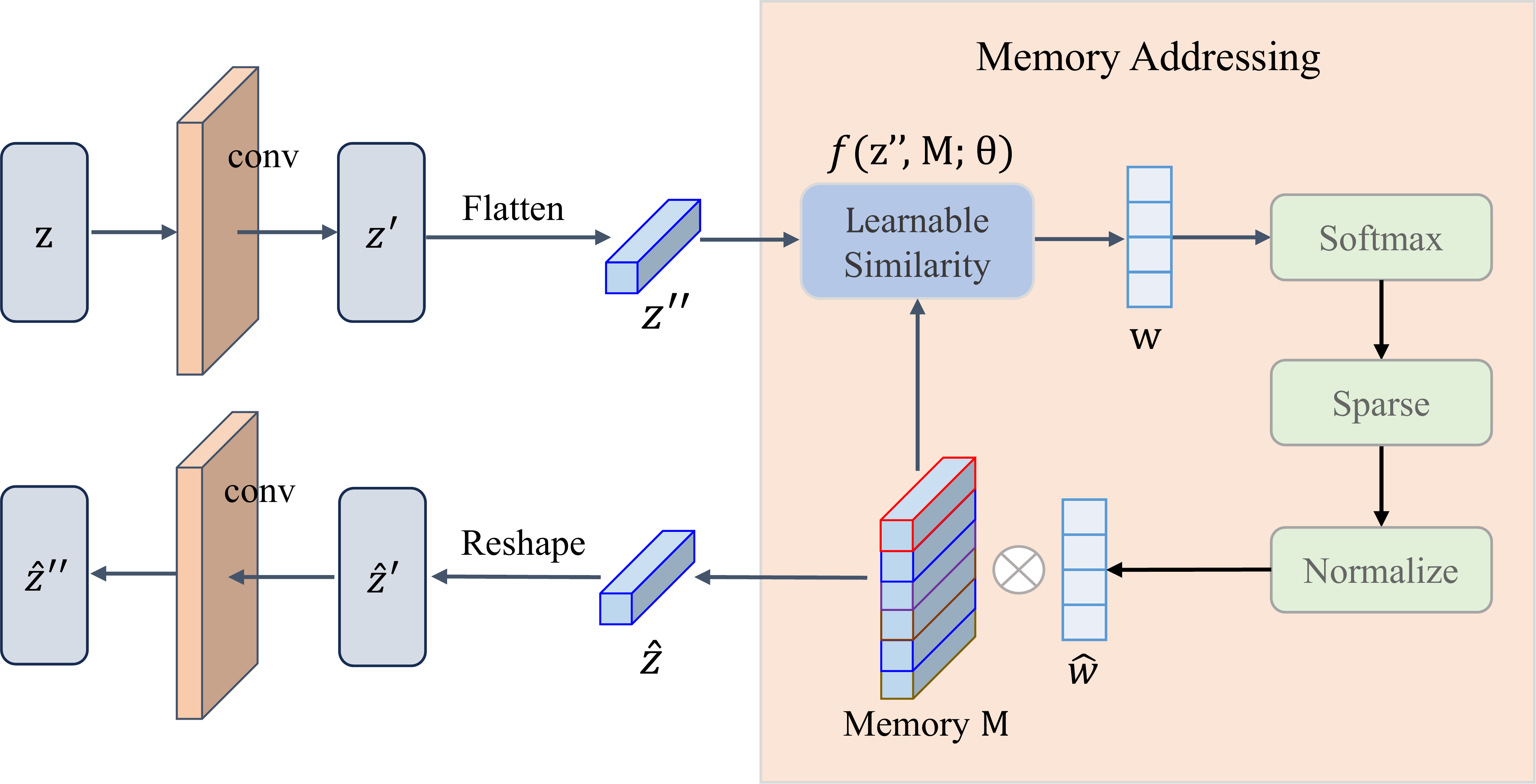}
    \caption{The memory module architecture. Given a feature vector, we first use a convolutional layer to reduce its dimensionality, and then find the most relevant codes in memory by an MLP network, based on which we can obtain the representations of input on memory.  Finally, such representation is input to the convolutional layer for reconstruction.}
    \label{memory}
\end{figure}

The Memory module, as in Fig. \ref{OverallFramework}, is proposed to remove the perturbations of vein samples by searching the most relevant latent codes to the given perturbed image from the memory.  Different from existing works \cite{2020Memorizing} \cite{chen2020mama} \cite{2020Learning} \cite{2021Memory}\cite{adhikarla2022memory}, we design a learnable metric to compute the difference between the input image latent code and the items in memory.  As shown in Fig. \ref{memory},  the proposed memory module includes two convolutional layers and a learnable memory dictionary.

Memory is a {learnable} matrix $M\in\mathbb{R}^{N\times d}$ containing $N$ real-valued vectors of fixed dimension $d$ {which are updated along with the model}. We denote the $i_{th}$ memory item as $m_i\in\mathbb{R}^{ d}$, i.e., the $i_{th}$ row of matrix $M$, where $\mathrm{i\in[0,N-1]}$. Given a query vector (i.e., an input image latent code) $z\in\mathbb{R}^{W\times H \times C}$, we perform a convolutional operation to obtain $z'\in\mathbb{R}^{W\times H \times c}$, which is flattened to a one-dimensional vector $z''$. For convenience, we assume that $d$ has the same dimension as $z''$. The memory module employs a soft $addressing $ vector $w\in\mathbb{R}^{1\times N}$ to obtain $\hat{z}$ by Eq. (\ref{8}):
\begin{equation}\hat{z}=wM=\sum_{i=1}^Nw_im_i, \label{8}\end{equation}
where $w$ is a row vector with non-negative elements with their sum equal to 1, and $w_i$ is the $i_{th}$ element of $w$. It is necessary to access memory $M$ to compute addressing weight $w$. The hyperparameter $N$ denotes the memory maximum capacity. 
The addressing weight vector $w$ is obtained based on $z''$ and $M$ by Eq. (\ref{9}):
\begin{equation}w_i=\frac{exp\left(s(z'',m_i)\right)}{\sum_{j=1}^Nexp\left(s(z'',m_j)\right)}, \label{9}\end{equation}

In existing works \cite{2020Memorizing} \cite{adhikarla2022memory}, the function $s(z , m_i)$ is defined as the cosine similarity between $z''$ and $m_i$ as Eq. (\ref{10}):
\begin{equation}s(z'',m_i)=\frac{z''m_i^T}{\|z''\|\|m_i\|}, \label{10}\end{equation}

As the cosine similarity is a handcrafted metric, it may not accurately describe the similarity of two vectors. To address this issue, we propose a learnable metric for measuring the difference between a latent code and memory items. In our model, the similarity between $z''$ and $m_i$ is computed through a multi-layer perception ($MLP$) in Eq. (\ref{11}). 
\begin{equation}s(z'',M)=f(z'',M;\theta), \label{11}\end{equation}
where $\theta$ is the learnable parameter in MLP $f$, detailed in Fig. \ref{MLP}:
\begin{figure}[h!]
    \centering
    \includegraphics[width=1\linewidth]{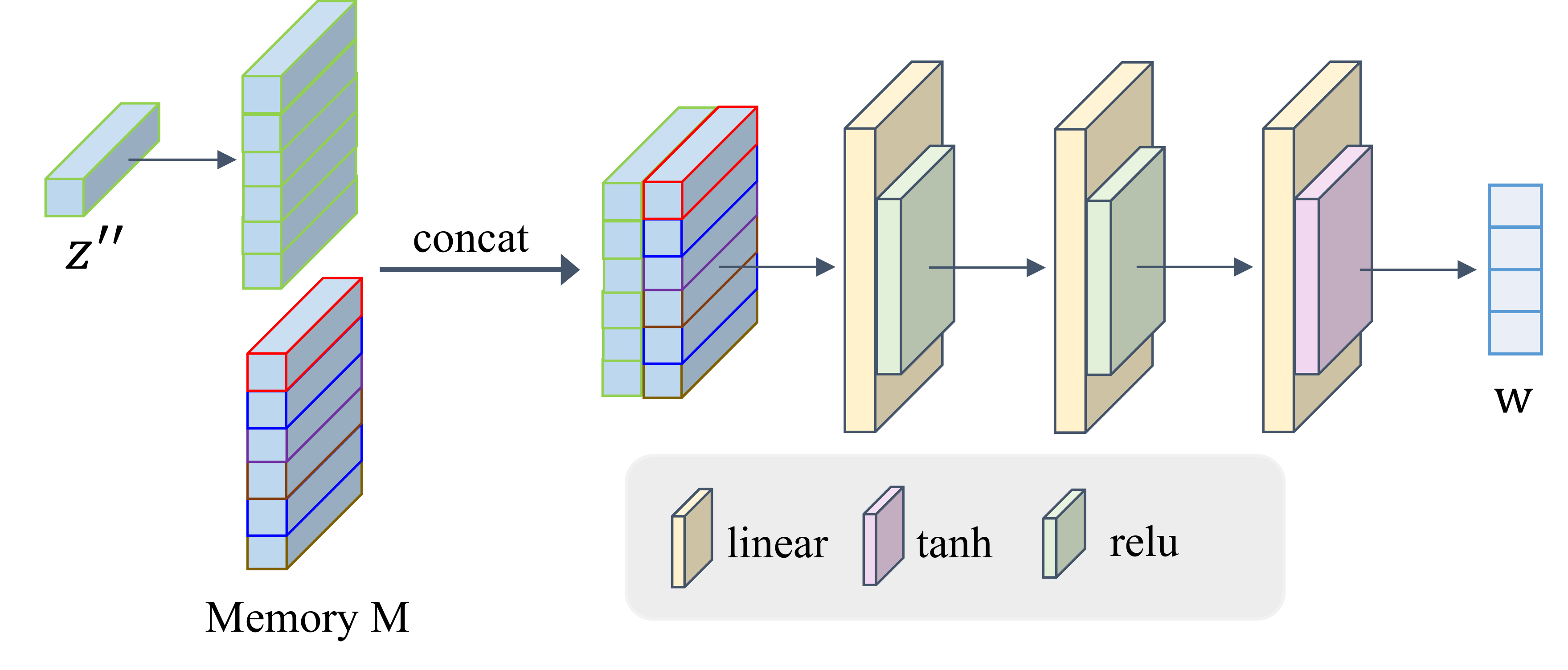}
    \caption{The MLP network architecture. Given an input vector $z''$, we duplicate it $N$ times to obtain $N$ vectors and combine them with memory $M$ (with size $N$). The resulting feature map passes then through the MLP's three linear layers. The first two linear layers are followed by a Relu activation function, while the last linear layer is followed by a Tanh activation function. The output $w\in\mathbb{R}^{1\times N}$ is finally  obtained.}
    \label{MLP}
\end{figure}

Employing a restricted number of normal patterns in the memory to reconstruct the image can help induce a large reconstruction error on an adversarial sample. In other words, the memory autoencoder with only $N$ memory items can not well reconstruct an adversarial sample, leading thereby to the purification of adversarial samples. It is possible, however, that some adversarial samples can be reconstructed well with a complex combination of the memory items based on a dense weight $w$. To achieve better purification, we propose a sparse addressing approach to reconstruct a sample by only a small number of memory items, as computed by Eq. (\ref{12}):

\begin{equation}\widehat{w}_i=normalize(\frac{max(w_i-\gamma,0)\cdot w_i}{|w_i-\gamma|+\alpha}), \label{12}\end{equation}
where $max(\cdot, 0)$ is the $ReLU$ activation, $\alpha$ is a very small positive number to prevent the denominator from being 0, and $\gamma$ is the weight threshold. In practice, similar to work \cite{2020Memorizing}, the threshold $\lambda$ is chosen as one of the values in set $[\frac{1}{N},\frac{3}{N}]$. The $normalize$ is a normalization function. After sparsification, we rewrite Eq. (\ref{8}) to Eq. (\ref{13}) as follows:
\begin{equation}\hat{z}=\widehat{w}M=\sum_{i=1}^N\widehat{w}_im_i, \label{13}\end{equation} 
where $\widehat{w}=[\widehat{w}_1,\widehat{w}_2,...\widehat{w}_N]$. Note that many values of elements in $\widehat{w}$  are equal to 0. As shown in Fig. \ref{memory}, the  $\hat{z}$ is subject to reshape to obtain the $\hat{z}^{'}\in\mathbb{R}^{W\times H \times c}$ and finally the output $\hat{z}^{''}\in\mathbb{R}^{W\times H \times C}$ is obtained by a convolution operation. Sparse addressing encourages the model to construct an image with fewer addressed memory items, to learn more informative representations of normal patterns in memory. This prevents the model from accurately reconstructing the perturbations with lots of dense addressing weights, allowing it thereby to filter adversarial perturbations.

\subsection{Patch-based Discriminator}
\begin{figure}[h!]
    \centering
    \includegraphics[width=0.9\linewidth]{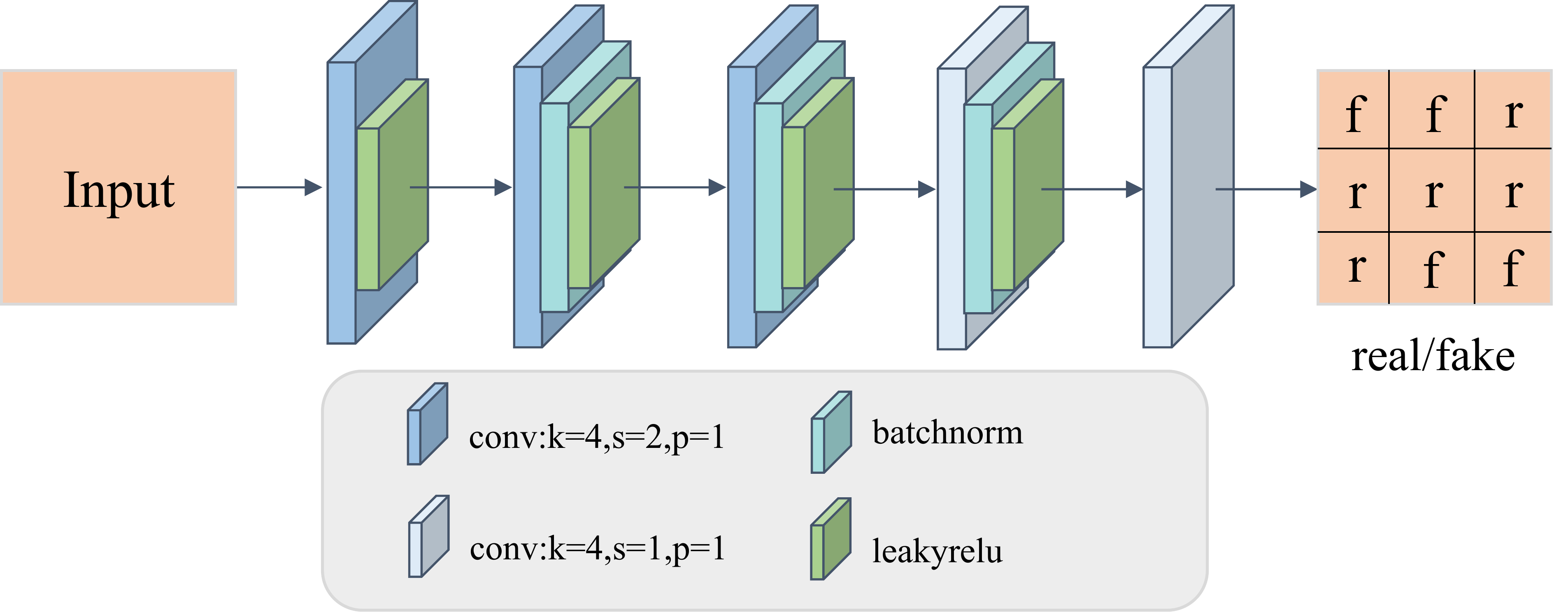}
    \caption{The architecture of the discriminator. The latter consists of five convolutional layers, where the first only is followed by a LeakyRelu activation function, while the middle three are followed by the batchnorm function and LeakyRelu. The last convolutional layer is used to output a matrix in which each value represents the evaluation value of a small region of the original image.}
    \label{disc}
\end{figure}

In this section, We propose, as shown in Fig. \ref{OverallFramework}, a patch-based discriminator to differentiate between real and reconstructed images, compared to the $L2$ pixel loss in \cite{2020Memorizing} \cite{adhikarla2022memory}, the patch-based discriminator \cite{2016Image} is capable of performing strong compression and retaining good perceptual quality in reconstructed images.
As shown in Fig.\ref{disc}, the discriminator learns to distinguish whether the patches are from real images or images generated by the corresponding generator.

The discriminator consists of five convolutional layers and three batch normalization layers. In the first four convolutional layers, $leakyrule$ is employed to extract the features. The middle three convolutional layers are followed by normalization layers. Given an input image, the discriminator predicts each of the input patches as real or fake (i.e. generated).       

\subsection{Loss Functions}
The per-pixel losses in \cite{2020Memorizing} \cite{adhikarla2022memory}  do not capture perceptual and semantic differences between output and original images. To address this limitation, we combine 
two perceptual losses, i.e. a feature reconstruction loss and an adversarial loss \cite{2016Image} with a pixel loss as a metric to differentiate between real and reconstructed images. Let $\{\mathbf{x}_l\}_{l=1}^L$  be a dataset with $L$ images and  ${{\Bar{x}}}_l$ denotes the reconstructed sample from original sample $x_l$. All losses are introduced as follows.

\textbf{$L1$ reconstruction loss.} Inspired by \cite{2016Image} which showed that the $L1$ distance instead of $L2$ encourages less blurring in reconstructed images, we use the $L1$ metric to measure the reconstruction loss by Eq. (\ref{14}):
\begin{equation}L_1(x,\Bar{x})=\parallel x^L-\Bar{x}^L\parallel_1, \label{14}\end{equation}

\textbf{Sparse loss:}  Similar to original MemAE \cite{2020Memorizing}, to further promote the sparsity of $\hat{{w}}$, we add a  sparse regularization term on memory, by minimizing the entropy of $\hat{\mathbf{w}}$, during the training process, which is computed by Eq. (\ref{15}):
\begin{equation}L_s(\widehat{w}^l)=\sum_{i=1}^L-\widehat{w}_i\cdot log(\widehat{w}_i), \label{15}\end{equation}
In fact, both Eq. (\ref{12}) and Eq. (\ref{15}) jointly promote the sparsity of the addressing weights  $\hat{{w}}$.

\begin{figure}[h!]
    \centering
    \includegraphics[width=0.8\linewidth]{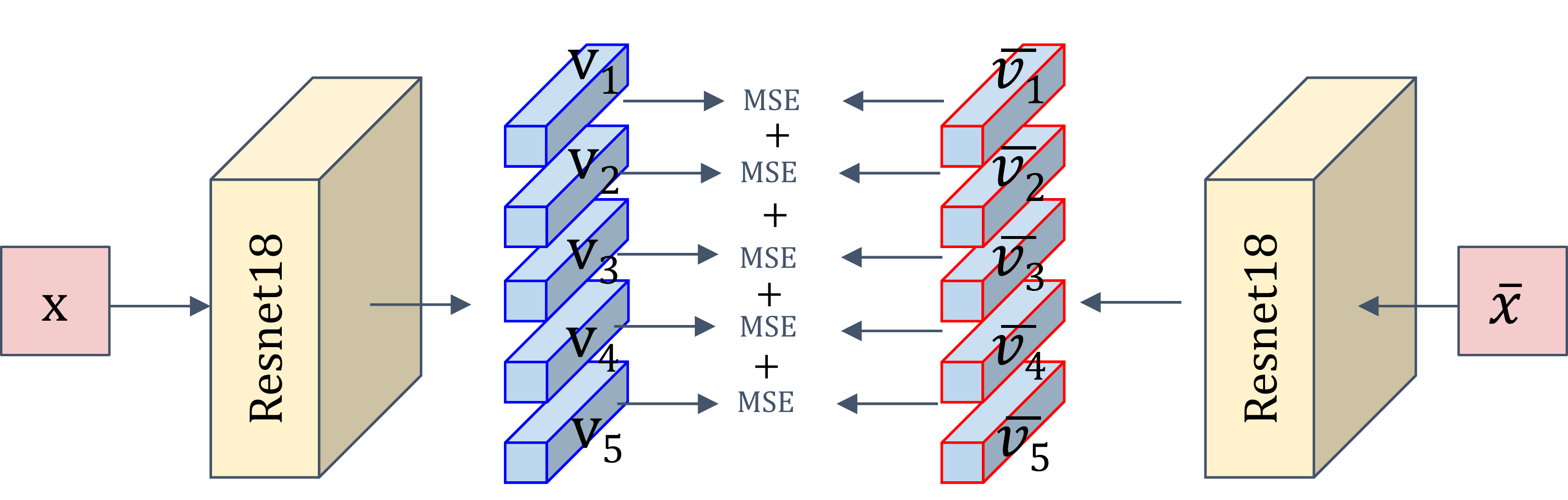}
    \caption{The original and reconstructed images are fed into resnet18 to obtain the perceptual loss.}
    \label{ploss}
\end{figure}

\textbf{Feature reconstruction loss:} Rather than exactly matching the pixels of the reconstructed image $\Bar{x}$ and the target image $x$, the perceptual loss computes the distance between the feature representations of the input image and the reconstructed image, which can improve the quality of the reconstructed image \cite{johnson2016perceptual}. First, we use a pre-trained Resnet18 network on Imagnet \cite{2012ImageNet} to extract feature representations from both the input image and the reconstructed image. Then, the $L2$ loss is employed to compute the distance between the two feature vectors. Let $\psi$ be pre-trained Resnet18 model. $\psi_i(x)$ is the feature map of the $i{th}$ layer of $\psi$ with an input image $x$. We obtain feature vector $v_i$ by flattening $\psi_i(x)$. Similarly, we can compute the  feature vector $\Bar{v}_i$ of the reconstructed image $\Bar{x}$ from  $\psi_(\Bar{x})$. The perceptual reconstruction loss  between the feature representations is defined by Eq. (\ref{16}):
\begin{equation}L_p(x,\Bar{x})=\frac1T\sum_{i=1}^T\left|\left|v_i-\Bar{v}_i\right|\right|_2^2, \label{16}\end{equation}
where $T$ is the number of feature layers ($T=5$). Fig. \ref{ploss} shows the computation process of the perceptual loss.

\textbf{Adversarial loss: } The classic GAN is difficult to train as the divergences that it typically minimizes are potentially not continuous w.r.t the generator's parameters. To address this problem, we employ the WGAN loss \cite{2017Wasserstein} and Hinge GAN loss \cite{lim2017geometric} to train our model, as defined by Eq. (\ref{17}) and Eq. (\ref{18}):

\begin{equation}
L_G=-E_{x \sim \mathbb{P}_{{x}}}[D(G(x))]
\label{17}
\end{equation}

\begin{equation}
\begin{aligned}
L_D&=E_{x \sim \mathbb{P}_{{x}}}[max(0,1-D(x))]\\
&+E_{x \sim \mathbb{P}_{{x}}}[max(0,1+D(G(x))]
\end{aligned}
\label{18}
\end{equation}

\begin{equation}L_{adv}=\beta L_G + L_D \label{19} \end{equation}
where $\mathbb{P}_{x}$ is the data distribution. Note that $G$ is the generator, namely the multi-scale memory AutoEncoder detailed in Section \ref{msmemae} and  $D$ is the discriminator, as shown in Fig. \ref{disc}. $\beta$ is an adaptive weight of the adversarial loss of $L_G$, as computed by Eq. (\ref{20}):
\begin{equation}\beta=\frac{\nabla_{G_\mathcal{L}}(L_1+L_p)}{\nabla_{G_\mathcal{L}}(L_{adv})+\sigma}\label{20}\end{equation}
where $L_p$ is the perceptual reconstruction loss and $L_{adv}$ is the adversarial loss, $\nabla_{G_\mathcal{L}}(\cdot)$ denotes the gradient of the input w.r.t the last layer $\mathcal{L}$ of decoder ${D}_t$ in Eq. (\ref{7}), and $\sigma=10^{-4}$ is used for stable training.

Finally, we combine  Eq. (\ref{14}), Eq. (\ref{15}), Eq. (\ref{16}), and Eq. (\ref{19}) to obtain the complete objective loss by Eq. (\ref{21}):

\begin{equation}
\begin{aligned}
L= L_1+L_p+\alpha L_s+L_{adv},
\end{aligned}
\label{21}
\end{equation}
where $\alpha$ is the weight of the information entropy loss. In our experiments, we experimentally set $\alpha$ to $0.0002$.

\subsection{Training and testing}
{The proposed MsMemoryGAN is trained in an end-to-end manner, i.e., the parameters of all the proposed modules are updated simultaneously by the gradient descent method. During the model training, the inputs are \textbf{normal} vein image samples, resulting in the pattern of normal samples stored all over the memory module.} The decoder in MsMemoryGAN uses a limited number of memory items to reconstruct normal samples by sparse addressing, which promotes efficient utilization of memory. 
Minimizing the reconstruction error encourages the memory to record the most representative patterns from the input normal samples. 
During testing, the normal patterns stored in the memory will be retrieved to represent the feature extracted by the encoder and the decoder will reconstruct original samples by using the represented features. As a result, the normal samples are reconstructed accurately. When taking an 
{adversarial} sample as the input, the normal patterns are also retrieved for reconstruction, as no perturbed items exist in the memory module, {causing the adversarial sample to be reconstructed back to the relatively normal sample.}
On the other hand, the perturbations of adversarial samples are usually not well reconstructed by retrieving patterns in the normal samples, thereby purifying the perturbations of the adversarial samples.

\section{Experiments and Results}
To evaluate the performance of our approach, we conducted extensive experiments on two public palm vein datasets, TJU\_PV dataset\cite{zhang2018palmprint} and PolyU\_MN dataset\cite{zhang2009online}, captured by contactless and contact devices at different times, respectively. First, black-box and white-box adversarial attack methods, namely, FGSM\cite{goodfellow2014explaining}, PGD\cite{madry2018towards}, SPSA\cite{uesato2018adversarial} were used to generate adversarial vein images by adding imperceptible perturbations. We then employed various defense models, namely MemAE\cite{2020Memorizing}, MemoryDefense\cite{adhikarla2022memory}, Magnet\cite{meng2017magnet}, DefenseGAN\cite{samangouei2018defense}, VeinGuard\cite{li2023transformer}, and DiffPure\cite{nie2022diffusion} to purify the perturbations. 
For {recognition}
the resulting cleaned images were input to state-of-the-art recognition approaches, namely Res2Net\cite{gao2019res2net}, Vit\cite{dosovitskiy2010image}, SwinTransformer\_v2\cite{liu2022swin}, FV\_CNN\cite{das2018convolutional}, PV-CNN\cite{9354642}, FVRAS-Net\cite{yang2020fvras}, and Lightweight-CNN\cite{shen2021finger}, as well as DefenseGAN\_ModelB\cite{samangouei2018defense}.

\begin{figure}[h!]
	\centering{
		\begin{minipage}[b]{0.22\linewidth}\centering
			\subfloat[]{\includegraphics[scale=0.12]{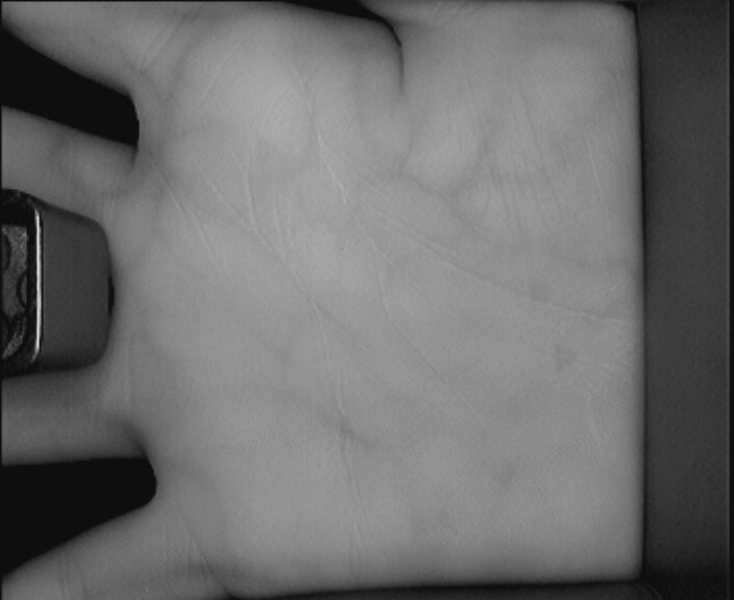}}\\
		\end{minipage}
		\begin{minipage}[b]{.22\linewidth}\centering
			\subfloat[]{\includegraphics[scale=0.13]{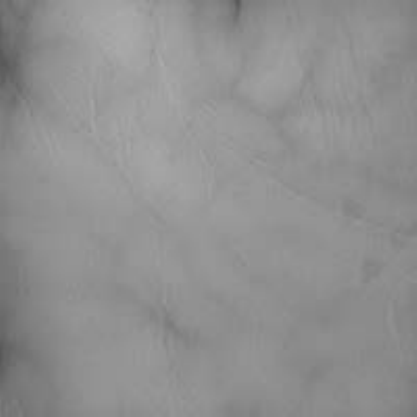}}\\
		\end{minipage}
	\begin{minipage}[b]{0.22\linewidth}\centering
			\subfloat[]{\includegraphics[scale=0.20]{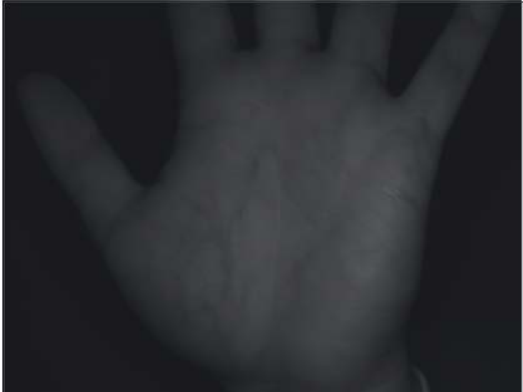}}\\
		\end{minipage}
		\begin{minipage}[b]{.22\linewidth}\centering
			\subfloat[]{\includegraphics[scale=0.21]{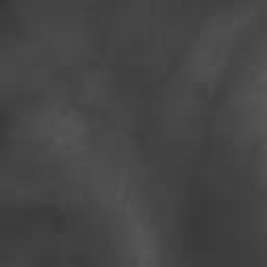}}\\
		\end{minipage}

	}
	\caption{Preprocessing results: (a) Original image from dataset A; (b) ROI from (a); (c) Original image from dataset B; (d) ROI from (c)}
	\label{oriandroi}
\end{figure}
\subsection{Datasets}
(1) \textbf{TJU\_PV}: TJU\_PV is a palmprint dataset collected at Tongji University \cite{zhang2018palmprint}, collected in a contactless way at intervals of about 2 months. The collected data are subject to variations such as illumination, translation, rotation, and scale. In each session, each volunteer provides two hands, each of which provides 10 images, resulting in 40 images for each volunteer (10 images $\times$ 2 sessions $\times$ 2 hands). For 300 volunteers, there is thus a total of 12,000 (10 images $\times$ 2 sessions $\times$ 2 hands $\times$ 300 volunteers) palmprint images. If we treat each hand as a category, we obtain a total of 600 categories with 20 images per category. As ROI (region of interest) images were extracted from the original palm-vein images, we directly employ them for model training and evaluation. An original palm-vein image and its normalized ROI are shown in Fig. \ref{oriandroi}(a) and Fig. \ref{oriandroi}(b).

(2) \textbf{PolyU M\_N}: The PolyU Multispectral Palmprint dataset \cite{zhang2009online}, collected at Hong Kong Polytechnic University, uses advanced multispectral imaging equipment for palmprint image acquisition, under blue, green, near-infrared (NIR), and red light illumination. As we focus on palm-vein recognition, only the images collected with NIR light are used in our experiments. The resulting palm-vein dataset comprises 6,000 palm-vein NIR images (250 subjects $\times$2 hands $\times$ 12 images) from 250 volunteers, with two hands for each volunteer, and each hand providing 12 images. As the images of each hand are treated as a class, we obtain a total of 500 classes, 12 images per class. Each image is subjected to ROI extraction and alignment by the method in \cite{qin2019iterative}. An original image and its normalized ROI are shown in Fig. \ref{oriandroi}(c) and Fig. \ref{oriandroi}(d). 
\begin{table*}[h!]
\setlength{\abovecaptionskip}{3pt} 
\setlength{\belowcaptionskip}{2pt} 
\renewcommand\arraystretch{1.5}
\centering
\caption{\centering Recognition accuracy of different recognition models using different defense methods on the TJU\_PV dataset under different strengths of FGSM and PGD white-box attacks.}
\scalebox{0.72}{
\begin{tabular}{ccccccccccc}
\toprule[1pt]
\multirow{1}{3cm}{\centering Method}  & \multirow{1}{*}{\centering Attack}  & \multirow{1}{*}{\centering Clean} & \multirow{1}{*}{\centering Adversarial} & \multirow{1}{*}{\centering MemAE\cite{2020Memorizing}} & \multirow{1}{*}{\centering MemoryDefense\cite{adhikarla2022memory}} & \multirow{1}{*}{\centering Magnet\cite{meng2017magnet}} & \multirow{1}{*}{\centering DefenseGAN\cite{samangouei2018defense}} & \multirow{1}{*}{\centering VeinGuard\cite{li2023transformer}} & \multirow{1}{*}{\centering DiffPure\cite{nie2022diffusion}} & \multirow{1}{*}{\centering \textbf{MsMemoryGAN}}  \\
\midrule[0.5pt]
\multirow{4}{3cm}{\centering FV\_CNN\cite{das2018convolutional}}  & FGSM,$L_\infty$,$\epsilon$=0.03  & \multirow{4}{*}{0.953} & 0.149  & 0.462 & 0.570 & 0.468 & 0.473 & 0.605 & 0.547 & \textbf{0.846} \\
                                         & FGSM,$L_\infty$,$\epsilon$=0.05  &                        & 0.022  & 0.499 & 0.556 & 0.450 & 0.407 & 0.575 & 0.564 & \textbf{0.791} \\
                                         & PGD,$L_\infty$,$\epsilon$=0.03   &                        & 0.786  & 0.469 & 0.579 & 0.777 & 0.357 & 0.632 & 0.498 & \textbf{0.840} \\
                                         & PGD,$L_\infty$,$\epsilon$=0.05   &                        & 0.651  & 0.475 & 0.575 & 0.677 & 0.352 & 0.581 & 0.493 & \textbf{0.835} \\
\midrule[0.5pt]
\multirow{4}{*}{\centering PV\_CNN\cite{9354642}}  & FGSM,$L_\infty$,$\epsilon$=0.03  & \multirow{4}{*}{0.998} & 0.882  & 0.849 & 0.755 & 0.904 & 0.770 & 0.843 & 0.772 & \textbf{0.911} \\
                                         & FGSM,$L_\infty$,$\epsilon$=0.05  &                        & 0.676  & 0.810 & 0.707 & 0.753 & 0.696 & 0.761 & 0.766 & \textbf{0.953} \\
                                         & PGD,$L_\infty$,$\epsilon$=0.03   &                        & 0.899  & 0.838 & 0.768 & 0.935 & 0.764 & 0.854 & 0.752 & \textbf{0.984} \\
                                         & PGD,$L_\infty$,$\epsilon$=0.05   &                        & 0.740  & 0.818 & 0.753 & 0.821 & 0.707 & 0.781 & 0.744 & \textbf{0.977} \\
\midrule[0.5pt]
\multirow{4}{*}{\centering Lightweight\_FVCNN\cite{shen2021finger}} & FGSM,$L_\infty$,$\epsilon$=0.03 & \multirow{4}{*}{0.955} & 0.691 & 0.668 & 0.576 & 0.890 & 0.635 & 0.784 & 0.672 & \textbf{0.909} \\
                                         & FGSM,$L_\infty$,$\epsilon$=0.05  &                        & 0.471  & 0.665 & 0.557 & 0.817 & 0.618 & 0.719 & 0.661 & \textbf{0.894} \\
                                         & PGD,$L_\infty$,$\epsilon$=0.03   &                        & 0.813  & 0.674 & 0.576 & 0.907 & 0.633 & 0.852 & 0.647 & \textbf{0.914} \\
                                         & PGD,$L_\infty$,$\epsilon$=0.05   &                        & 0.657  & 0.671 & 0.578 & 0.887 & 0.618 & 0.769 & 0.642 & \textbf{0.915} \\
\midrule[0.5pt]
\multirow{4}{*}{\centering FVRAS\_Net\cite{yang2020fvras}}  & FGSM,$L_\infty$,$\epsilon$=0.03 & \multirow{4}{*}{0.997} & 0.673 & 0.749 & 0.582 & 0.838 & 0.673 & 0.753 & 0.734 & \textbf{0.966} \\
                                         & FGSM,$L_\infty$,$\epsilon$=0.05  &                        & 0.407  & 0.675 & 0.409 & 0.620 & 0.609 & 0.691 & 0.719 & \textbf{0.938} \\
                                         & PGD,$L_\infty$,$\epsilon$=0.03   &                        & 0.789  & 0.762 & 0.661 & 0.946 & 0.677 & 0.803 & 0.724 & \textbf{0.977} \\
                                         & PGD,$L_\infty$,$\epsilon$=0.05   &                        & 0.553  & 0.738 & 0.637 & 0.891 & 0.646 & 0.772 & 0.706& \textbf{0.971} \\
\midrule[0.5pt]
\multirow{4}{*}{\centering DefenseGAN\_ModelB\cite{samangouei2018defense}} & FGSM,$L_\infty$,$\epsilon$=0.03   & \multirow{4}{*}{0.977} & 0.809 & 0.709 & 0.594 & 0.929 & 0.727 & 0.836 & 0.782 & \textbf{0.939} \\
                                         & FGSM,$L_\infty$,$\epsilon$=0.05  &                        & 0.685  & 0.701 & 0.572 & 0.902 & 0.719 & 0.714 & 0.770 & \textbf{0.930} \\
                                         & PGD,$L_\infty$,$\epsilon$=0.03   &                        & 0.833  & 0.709 & 0.599 & 0.939 & 0.723 & 0.847 & 0.771 & \textbf{0.944} \\
                                         & PGD,$L_\infty$,$\epsilon$=0.05   &                        & 0.687  & 0.712 & 0.569 & 0.936 & 0.714 & 0.735 & 0.762 & \textbf{0.944} \\
\midrule[0.5pt]
\multirow{4}{*}{Res2Net50\cite{gao2019res2net}} & FGSM,$L_\infty$,$\epsilon$=0.01   & \multirow{4}{*}{0.978} & 0.880 & 0.845 & 0.806 & 0.928 & 0.789 & 0.879 & 0.662 & \textbf{0.943} \\
                           & FGSM,$L_\infty$,$\epsilon$=0.015  &                        & 0.768 & 0.735 & 0.801 & 0.884 & 0.766 & 0.865 & 0.668 & \textbf{0.932} \\
                           & PGD,$L_\infty$,$\epsilon$=0.01    &                        & 0.857 & 0.830 & 0.806 & 0.924 & 0.777 & 0.861 & 0.638 & \textbf{0.942} \\
                           & PGD,$L_\infty$,$\epsilon$=0.015   &                        & 0.752 & 0.819 & 0.803 & 0.871 & 0.744 & 0.823 & 0.631 & \textbf{0.933} \\
\midrule[0.5pt]
\multirow{4}{*}{Vit\cite{dosovitskiy2010image}}       & FGSM,$L_\infty$,$\epsilon$=0.01   & \multirow{4}{*}{0.973} & 0.650 & 0.627 & 0.706 & 0.756 & 0.650 & 0.737 & 0.521 & \textbf{0.909} \\
                           & FGSM,$L_\infty$,$\epsilon$=0.015  &                        & 0.401 & 0.513 & 0.606 & 0.571 & 0.611 & 0.712 & 0.508 & \textbf{0.903} \\
                           & PGD,$L_\infty$,$\epsilon$=0.01    &                        & 0.575 & 0.621 & 0.705 & 0.724 & 0.619 & 0.704 & 0.524 & \textbf{0.906} \\
                           & PGD,$L_\infty$,$\epsilon$=0.015   &                        & 0.318 & 0.511 & 0.607 & 0.549 & 0.694 & 0.659 & 0.498 & \textbf{0.906} \\
\midrule[0.5pt]
\multirow{4}{*}{SwinTransformer\_v2\cite{liu2022swin}}    & FGSM,$L_\infty$,$\epsilon$=0.01   & \multirow{4}{*}{0.997} & 0.564 & 0.879 & 0.776 & 0.802 & 0.767 & 0.897 & 0.762 & \textbf{0.980} \\
                           & FGSM,$L_\infty$,$\epsilon$=0.015  &                        & 0.261 & 0.864 & 0.775 & 0.646 & 0.732 & 0.806 & 0.741 & \textbf{0.971} \\
                           & PGD,$L_\infty$,$\epsilon$=0.01    &                        & 0.353 & 0.877 & 0.776 & 0.755 & 0.764 & 0.884 & 0.744 & \textbf{{0.981}} \\
                           & PGD,$L_\infty$,$\epsilon$=0.015   &                        & 0.102 & 0.866 & 0.776 & 0.545 & 0.729 & 0.878 & 0.723 & \textbf{0.978} \\
\midrule[0.5pt]
\multicolumn{2}{c}{Average / Gain} \vspace{5pt} \rule{0pt}{15pt} & 0.979 & 0.605 & 0.708 / 0.103 & 0.660 / {0.055} & 0.789 / 0.184 & 0.661 / 0.056 & 0.769 / 0.164 & 0.667 / 0.062 & \textbf{0.927} / \textbf{\textcolor{green}{0.322}} \\
\bottomrule[1pt]
\end{tabular}
}
\label{tab1}
\end{table*}

\begin{figure}[h!]
    \centering
    \includegraphics[width=0.8\linewidth]{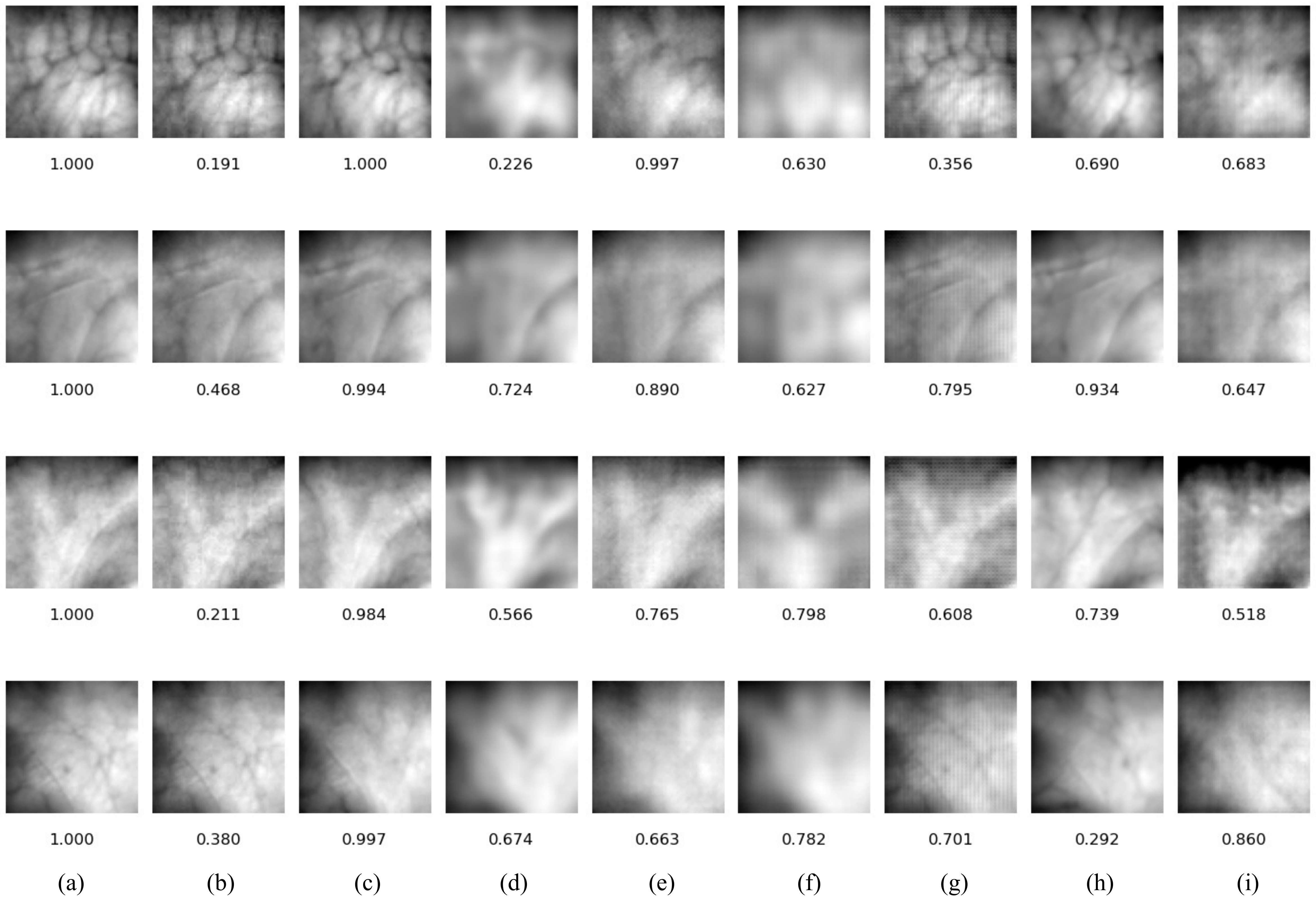}
    \caption{Reconstruction results of various on the TJU\_PV dataset: (a) Original image; (b) Adversarial image of (a)  obtained by FGSM attack; {(c) Reconstructed image of (b) using \textbf{MsMemoryGAN};} (d) Reconstructed image of (b) using MemAE; (e) Reconstructed image of (b) using VeinGuard; (f) Reconstructed image of (b) using MemoryDefense; (g) Reconstructed image of (b) using Magnet; (h) Reconstructed image of (b) using DiffPure; (i) Reconstructed image of (b) using DefenseGAN. We list the confidence scores calculated by the Vit model at the bottom of each image}
    \label{visual}
\end{figure}

\subsection{Experimental Settings}
First, we divide each dataset into training and test sets. Dataset TJU\_PV comprises 12,000 images from 600 classes, with each class providing 20 images. For each class, We use 15 images for training and remain 5 for testing. This generates 9,000 images (600 classes $\times$ 15 images) in the training set and 3,000 images (600 classes $\times$ 5 images) in the test set. Similarly, from each hand, 8 images are considered for training and 4 for testing in dataset PolyU\_MN, generating thereby a training set with 4000 images (500 classes $\times$ 8 images) and a test set with 2000 images (500 classes $\times$ 4 images). 
Then, Eight {recognition models}, 
namely Res2Net\cite{gao2019res2net}, Vit\cite{dosovitskiy2010image}, SwinTransformer\_v2\cite{liu2022swin}, FV\_CNN\cite{das2018convolutional}, PV\_CNN\cite{9354642}, FVRAS\_Net\cite{yang2020fvras}, Lightweight\_CNN\cite{shen2021finger}, and DefenseGAN\_ModelB \cite{samangouei2018defense}, are trained on normal images in the training sets. 
Attack methods, FGSM\cite{goodfellow2014explaining}, PGD\cite{madry2018towards}, and SPSA \cite{uesato2018adversarial} are applied to attack test images. Various defense approaches, namely MemAE\cite{2020Memorizing}, MemoryDefense\cite{adhikarla2022memory}, Magnet \cite{meng2017magnet}, DefenseGAN \cite{samangouei2018defense}, VeinGuard \cite{li2023transformer}, and DiffPure \cite{nie2022diffusion}, as well as our approach, are employed to remove the perturbations. Finally, for comparison, We evaluate the accuracy performance of all recognition models on originally normal (unperturbed) samples, adversarial samples, and purified samples. 
{ Noted that most biometric recognition systems require an enrolment process. Therefore, all recognition methods are trained to extract the input features and the similarity of registrations and test images is computed for recognition. The samples in the training set are treated as registered samples and the samples in the test set are matched for similarity with them.}  

For our approaches, hyperparameter $N$ is the memory size. The performance of the memory module is not sensitive to the size of $N$ for different datasets \cite{2020Memorizing}. Ideally, a large $N$ is suitable for all datasets. We set it to 1000 for both the TJU\_PV and PolyU\_MN datasets. The sparsity threshold $\gamma$ in Eq. (\ref{7}) is set to $\frac{1}{N}$. 
{ For model training, the adaptive moment estimation with weight decay (AdamW) \cite{5a260c8617c44a4ba8a3277c} optimizer is employed to model parameter optimization. The initial learning rate is set to 0.001, it will eventually reduce to 0.0001 with a cosine scheduler which has a warmup epoch number of 10, the weight decay is 0.05, and the minibatch size to 60 and 40 for TJU\_PV database and VERA\_PV database respectively. The maximum number of training epochs is set to 1000. All experiments are implemented on PyTorch with an NVIDIA Tesla A100 80G GPU. }

\subsection{Visual assessment}
In this section, we visually evaluate the performance of the various defense methods. Specifically, we performed FGSM attacks \cite{goodfellow2014explaining} with the Vit \cite{dosovitskiy2010image} on the test samples to generate adversarial samples. {The resulting images were input into the seven defense models
as well as our MsMemoryGAN to obtain the reconstructed images (purified images)}. We compute the probability of samples belonging to a given class before and after reconstruction. The reconstructed images by these defense approaches and their corresponding prediction scores are shown in Fig. \ref{visual}. To facilitate comparison, we also show the original vein image. 

Comparing Fig. \ref{visual}(a) and Fig. \ref{visual}(b), it is hard to differentiate them visually but the accuracy is degraded significantly, which is consistent with the results in Fig. \ref{attacksample}.  From Fig. \ref{visual}(a) and Fig. \ref{visual}(c), we observe that our MsMemoryGAN reconstructs high-quality images, as the purified samples generated by MsMemoryGAN have highly consistent distribution with the original images. For example, the details in the original images are almost all retained in the purified images, with a high resolution (Fig. \ref{visual}(c)). At the same time, the purified images still have high confidence scores, which implies that our MsMemoryGAN is capable of filtering the perturbation and defending recognition models against adversarial attacks. By contrast, there are large differences between the original images and purified images generated by the other approaches. Specifically, MemAE, MemoryDefense, Magnet, and VeinGuard generate blurred reconstructed images (Fig. \ref{visual}(d), Fig. \ref{visual}(f), Fig. \ref{visual}(g), and Fig. \ref{visual}(h)), where some vein textures are missing. Comparably, DiffPure achieves better reconstruction, but the reconstructed image (Fig. \ref{visual}(h)) still shows a different vein distribution w.r.t the original image (Fig. \ref{visual}(a)). Overall, MsMemoryGAN can reconstruct high-quality images while maintaining high confidence scores for the genuine class and outperforming the other approaches.

\begin{table*}[h!]
\setlength{\abovecaptionskip}{3pt} 
\setlength{\belowcaptionskip}{2pt} 
\renewcommand\arraystretch{1.5}
\centering
\caption{\centering Recognition accuracy of different recognition models using different defense methods on PolyU M\_N dataset under FGSM and PGD white-box attacks.}
\scalebox{0.72}{
\begin{tabular}{ccccccccccc}
\toprule[1pt]
\multirow{1}{3cm}{\centering Method}  & \multirow{1}{*}{\centering Attack}  & \multirow{1}{*}{\centering Clean} & \multirow{1}{*}{\centering Adversarial} & \multirow{1}{*}{\centering MemAE\cite{2020Memorizing}} & \multirow{1}{*}{\centering MemoryDefense\cite{adhikarla2022memory}} & \multirow{1}{*}{\centering Magnet\cite{meng2017magnet}} & \multirow{1}{*}{\centering DefenseGAN\cite{samangouei2018defense}} & \multirow{1}{*}{\centering VeinGuard\cite{li2023transformer}} & \multirow{1}{*}{\centering DiffPure\cite{nie2022diffusion}} & \multirow{1}{*}{\centering \textbf{MsMemoryGAN}}  \\
\toprule[1pt]
\multirow{2}{3cm}{\centering FV\_CNN\cite{das2018convolutional}}  & FGSM,$L_\infty$,$\epsilon$=0.3  & \multirow{2}{*}{0.976} & 0.005 & 0.460 & 0.516 & 0.405 & 0.583 & 0.631 & 0.607 & \textbf{0.780} \\
                                         & PGD,$L_\infty$,$\epsilon$=0.3   &                        & 0.272 & 0.550 & 0.534 & 0.488 & 0.594 & 0.617 & 0.864 & \textbf{0.867} \\
\midrule[0.5pt]
\multirow{2}{*}{\centering PV\_CNN\cite{9354642}}  & FGSM,$L_\infty$,$\epsilon$=0.3  & \multirow{2}{*}{0.993} & 0.176 & 0.581 & 0.651 & 0.582 & 0.496 & 0.896 & 0.905 & \textbf{0.909} \\
                                         & PGD,$L_\infty$,$\epsilon$=0.3   &                        & 0.528 & 0.627 & 0.734 & 0.928 & 0.515 & 0.834 & \textbf{0.972} & 0.967 \\
\midrule[0.5pt]
\multirow{2}{*}{\centering Lightweight\_FVCNN\cite{shen2021finger}}    & FGSM,$L_\infty$,$\epsilon$=0.3 & \multirow{2}{*}{0.885} & 0.168 & 0.320 & 0.459 & 0.665 & 0.429 & 0.528 & 0.769 & \textbf{0.807} \\
                                         & PGD,$L_\infty$,$\epsilon$=0.3   &                        & 0.212 & 0.310 & 0.528 & 0.626 & 0.512 & 0.649 & 0.814 & \textbf{0.848} \\
\midrule[0.5pt]
\multirow{2}{*}{\centering FVRAS\_Net\cite{yang2020fvras}}  & FGSM,$L_\infty$,$\epsilon$=0.3 & \multirow{2}{*}{0.992} & 0.714 & 0.879 & 0.581 & 0.788 & 0.367 & 0.594 & 0.916 & \textbf{0.961} \\
                                         & PGD,$L_\infty$,$\epsilon$=0.3   &                        & 0.880 & 0.883 & 0.662 & 0.920 & 0.762 & 0.853 & 0.957 & \textbf{0.973} \\
\midrule[0.5pt]
\multirow{2}{*}{\centering DefenseGAN\_ModelB\cite{samangouei2018defense}} & FGSM,$L_\infty$,$\epsilon$=0.3   & \multirow{2}{*}{0.914} & 0.729 & 0.778 & 0.508 & 0.757 & 0.585 & 0.804 & \textbf{0.886} & 0.881 \\
                                         & PGD,$L_\infty$,$\epsilon$=0.3   &                        & 0.768 & 0.768 & 0.549 & 0.791 & 0.553 & 0.828 & 0.887 & \textbf{0.906} \\                           
\midrule[0.5pt]
\multirow{2}{*}{Res2Net50\cite{gao2019res2net}} & FGSM,$L_\infty$,$\epsilon$=0.2   & \multirow{2}{*}{0.993} & 0.567 & 0.679 & 0.759 & 0.640 & 0.752 & 0.815 & 0.960 & \textbf{0.961} \\
                           & PGD,$L_\infty$,$\epsilon$=0.2    &                        & 0.029 & 0.650 & 0.785 & 0.547 & 0.738 & 0.838 & 0.966 & \textbf{0.968} \\
\midrule[0.5pt]
\multirow{2}{*}{Vit\cite{dosovitskiy2010image}}       & FGSM,$L_\infty$,$\epsilon$=0.2   & \multirow{2}{*}{0.900} & 0.095 & 0.695 & 0.494 & 0.497 & 0.440 & 0.637 & 0.754 & \textbf{0.890} \\
                           & PGD,$L_\infty$,$\epsilon$=0.2    &                        & 0.623 & 0.715 & 0.550 & 0.641 & 0.444 & 0.649 & 0.866 & \textbf{0.890} \\
\midrule[0.5pt]
\multirow{2}{*}{SwinTransformer\_v2\cite{liu2022swin}}    & FGSM,$L_\infty$,$\epsilon$=0.2   & \multirow{2}{*}{0.991} & 0.240 & 0.877 & 0.767 & 0.663 & 0.706 & 0.826 & 0.947 & \textbf{0.974} \\
                           & PGD,$L_\infty$,$\epsilon$=0.2    &                        & 0.220 & 0.891 & 0.840 & 0.645 & 0.716 & 0.807 & 0.972 & \textbf{{0.982}} \\
\midrule[0.5pt]
\multicolumn{2}{c}{Average / Gain} \vspace{5pt} \rule{0pt}{15pt} & 0.956 & 0.389 & 0.666 / 0.277 & 0.620 / 0.231 & 0.661 / 0.272 & 0.575 / {0.186} & 0.738 / 0.349 & 0.878 / 0.489 & \textbf{0.910} / \textbf{\textcolor{green}{0.521}} \\ 
\bottomrule[1pt]
\end{tabular}
}
\label{tab2}
\end{table*}

\subsection{Quantitative Assessment}
\subsubsection{White Box Attack Results}
To verify the defense performance of MsMemoryGAN, we report the experimental results of various recognition models under white-box attacks, where the attacker accesses full or partial knowledge of the target model such as the architecture, parameters, inputs, outputs, and even training data to achieve better attacks. White-box attacks usually have a high success rate in attacking recognition systems. 
In our experiments, we apply state-of-the-art white-box attack models, i.e. FGSM \cite{goodfellow2014explaining} and PGD \cite{madry2018towards} with parameter $\epsilon$, which is the maximum attack intensity. A larger $\epsilon$ results in a stronger attack, but with visible perturbations in the test samples. A smaller $\epsilon$, by contrast, may fail to attack the target model. 
To effectively test our defense method, we consider different $\epsilon$ values, 0.03 and 0.05 for the TJU\_PV dataset, and 0.3 and  0.2 for the PolyU M\_N dataset. The resulting purified samples are fed to the eight recognition models. The experimental results of various models with original, adversarial, and purified images on the two datasets are shown in Tab. \ref{tab1} and Tab. \ref{tab2}.

From Table \ref{tab1} and Table \ref{tab2}, {we observe that all recognition models achieve more than 95\% accuracy on the TJU\_PV dataset and more than 88\% accuracy on PolyU M\_N.} Their accuracy, however, is significantly reduced after the FGSM \cite{goodfellow2014explaining} and PGD \cite{madry2018towards} attacks, which demonstrates that the DL-based vein recognition systems are prone to adversarial attacks. This conclusion is consistent with the findings in \cite{goodfellow2014explaining}. After removing the adversarial perturbations from the test samples by existing defense approaches and our MsMemoryGAN, the accuracy is significantly improved, which means that the defense approaches are capable of defending the vein recognition system against adversarial attacks.

\begin{table*}[h!]
\setlength{\abovecaptionskip}{3pt} 
\setlength{\belowcaptionskip}{2pt} 
\renewcommand\arraystretch{2}
\centering
\caption{\centering Recognition accuracy of various recognition models using different defense strategies on TJU\_PV dataset under SPSA black-box attack.}
\scalebox{0.72}{
\begin{tabular}{ccccccccccc}
\toprule[1pt]
\multirow{1}{3cm}{\centering Method}  & \multirow{1}{*}{\centering Attack}  & \multirow{1}{*}{\centering Clean} & \multirow{1}{*}{\centering Adversarial} & \multirow{1}{*}{\centering MemAE\cite{2020Memorizing}} & \multirow{1}{*}{\centering MemoryDefense\cite{adhikarla2022memory}} & \multirow{1}{*}{\centering Magnet\cite{meng2017magnet}} & \multirow{1}{*}{\centering DefenseGAN\cite{samangouei2018defense}} & \multirow{1}{*}{\centering VeinGuard\cite{li2023transformer}} & \multirow{1}{*}{\centering DiffPure\cite{nie2022diffusion}} & \multirow{1}{*}{\centering \textbf{MsMemoryGAN}}  \\
\toprule[1pt]
\multirow{1}{3cm}{\centering FV\_CNN\cite{das2018convolutional}} & \multirow{1}{*}{\centering SPSA,$L_\infty$,$\epsilon$=0.1} & \multirow{1}{*}{\centering 0.953} & \multirow{1}{*}{\centering 0.718} & \multirow{1}{*}{\centering 0.569} & \multirow{1}{*}{\centering 0.659} & \multirow{1}{*}{\centering 0.668} & \multirow{1}{*}{\centering 0.450} & \multirow{1}{*}{\centering 0.646} & \multirow{1}{*}{\centering 0.514} & \multirow{1}{*}{\centering \textbf{0.806}} \\
\midrule[0.5pt]
\multirow{1}{*}{\centering PV\_CNN\cite{9354642}} & \multirow{1}{*}{\centering SPSA,$L_\infty$,$\epsilon$=0.2} & \multirow{1}{*}{\centering 0.998} & \multirow{1}{*}{\centering 0.717} & \multirow{1}{*}{\centering 0.756} & \multirow{1}{*}{\centering 0.649} & \multirow{1}{*}{\centering 0.857} & \multirow{1}{*}{\centering 0.709} & \multirow{1}{*}{\centering 0.884} & \multirow{1}{*}{\centering 0.758} & \multirow{1}{*}{\centering \textbf{0.953}} \\
\midrule[0.5pt]
\multirow{1}{*}{\centering Lightweight\_FVCNN\cite{shen2021finger}} & \multirow{1}{*}{\centering SPSA,$L_\infty$,$\epsilon$=0.1} & \multirow{1}{*}{\centering 0.955} & \multirow{1}{*}{\centering 0.741} & \multirow{1}{*}{\centering 0.651} & \multirow{1}{*}{\centering 0.662} & \multirow{1}{*}{\centering 0.879} & \multirow{1}{*}{\centering 0.594} & \multirow{1}{*}{\centering 0.834} & \multirow{1}{*}{\centering 0.651} & \multirow{1}{*}{\centering \textbf{0.900}} \\
\midrule[0.5pt]
\multirow{1}{*}{\centering FVRAS\_Net\cite{yang2020fvras}}  & \multirow{1}{*}{\centering SPSA,$L_\infty$,$\epsilon$=0.2} & \multirow{1}{*}{\centering 0.997} & \multirow{1}{*}{\centering 0.371} & \multirow{1}{*}{\centering 0.633} & \multirow{1}{*}{\centering 0.672} & \multirow{1}{*}{\centering 0.683} & \multirow{1}{*}{\centering 0.543} & \multirow{1}{*}{\centering 0.756} & \multirow{1}{*}{\centering 0.677} & \multirow{1}{*}{\centering \textbf{0.919}} \\
\midrule[0.5pt]
\multirow{1}{*}{\centering DefenseGAN\_ModelB\cite{samangouei2018defense}} & \multirow{1}{*}{\centering SPSA,$L_\infty$,$\epsilon$=0.1}  & \multirow{1}{*}{\centering 0.977} & \multirow{1}{*}{\centering 0.770} & \multirow{1}{*}{\centering 0.693} & \multirow{1}{*}{\centering 0.681} & \multirow{1}{*}{\centering 0.839} & \multirow{1}{*}{\centering 0.710} & \multirow{1}{*}{\centering 0.842} & \multirow{1}{*}{\centering 0.768} & \multirow{1}{*}{\centering \textbf{0.941}} \\
\midrule[0.5pt]
\multirow{1}{*}{Res2Net50\cite{gao2019res2net}} & \multirow{1}{*}{\centering SPSA,$L_\infty$,$\epsilon$=0.1} & \multirow{1}{*}{\centering 0.978} & \multirow{1}{*}{\centering 0.739} & \multirow{1}{*}{\centering 0.778} & \multirow{1}{*}{\centering 0.706} & \multirow{1}{*}{\centering 0.856} & \multirow{1}{*}{\centering 0.700} & \multirow{1}{*}{\centering 0.861} & \multirow{1}{*}{\centering 0.614} & \multirow{1}{*}{\centering \textbf{0.917}} \\
\midrule[0.5pt]
\multirow{1}{*}{Vit\cite{dosovitskiy2010image}}       & \multirow{1}{*}{\centering SPSA,$L_\infty$,$\epsilon$=0.1} & \multirow{1}{*}{\centering 0.973} & \multirow{1}{*}{\centering 0.356} & \multirow{1}{*}{\centering 0.512} & \multirow{1}{*}{\centering 0.575} & \multirow{1}{*}{\centering 0.570} & \multirow{1}{*}{\centering 0.408} & \multirow{1}{*}{\centering 0.623} & \multirow{1}{*}{\centering 0.520} & \multirow{1}{*}{\centering \textbf{0.881}} \\
\midrule[0.5pt]
\multirow{1}{*}{SwinTransformer\_v2\cite{liu2022swin}}    & \multirow{1}{*}{\centering SPSA,$L_\infty$,$\epsilon$=0.1} & \multirow{1}{*}{\centering 0.997} & \multirow{1}{*}{\centering 0.761} & \multirow{1}{*}{\centering 0.883} & \multirow{1}{*}{\centering 0.704} & \multirow{1}{*}{\centering 0.899} & \multirow{1}{*}{\centering 0.786} & \multirow{1}{*}{\centering 0.875} & \multirow{1}{*}{\centering 0.768} & \multirow{1}{*}{\centering \textbf{{0.977}}} \\
\midrule[0.5pt]
\multicolumn{2}{c}{Average / Gain} \vspace{5pt} \rule{0pt}{15pt} & 0.979 & 0.647 & 0.684 / 0.037 & 0.664 / 0.017 & 0.781 / 0.134 & 0.613 / {-0.034} & 0.790 / 0.143 & 0.659 / 0.012 & \textbf{0.912} / \textbf{\textcolor{green}{0.265}} \\ 
\bottomrule[1pt]
\end{tabular}
}
\label{tab3}
\end{table*}

\begin{table*}[h!]
\setlength{\abovecaptionskip}{3pt} 
\setlength{\belowcaptionskip}{2pt} 
\renewcommand\arraystretch{2}
\centering
\caption{\centering Recognition accuracy of various recognition models using different defense strategies on PolyU M\_N dataset under SPSA black box attack.}
\scalebox{0.72}{
\begin{tabular}{ccccccccccc}
\toprule[1pt]
\multirow{1}{3cm}{\centering Model}  & \multirow{1}{*}{\centering Attack}  & \multirow{1}{*}{\centering Clean} & \multirow{1}{*}{\centering Adversarial} & \multirow{1}{*}{\centering MemAE\cite{2020Memorizing}} & \multirow{1}{*}{\centering MemoryDefense\cite{adhikarla2022memory}} & \multirow{1}{*}{\centering Magnet\cite{meng2017magnet}} & \multirow{1}{*}{\centering DefenseGAN\cite{samangouei2018defense}} & \multirow{1}{*}{\centering VeinGuard\cite{li2023transformer}} & \multirow{1}{*}{\centering DiffPure\cite{nie2022diffusion}} & \multirow{1}{*}{\centering \textbf{MsMemoryGAN}}  \\
\toprule[1pt]
\multirow{1}{3cm}{\centering FV\_CNN\cite{das2018convolutional}} & \multirow{1}{*}{\centering SPSA,$L_\infty$,$\epsilon$=0.3} & \multirow{1}{*}{0.976} & \multirow{1}{*}{\centering 0.132} & \multirow{1}{*}{\centering 0.392} & \multirow{1}{*}{\centering 0.435} & \multirow{1}{*}{\centering 0.245} & \multirow{1}{*}{\centering 0.598} & \multirow{1}{*}{\centering 0.693} & \multirow{1}{*}{\centering 0.724} & \multirow{1}{*}{\centering \textbf{0.834}} \\
\midrule[0.5pt]
\multirow{1}{*}{\centering PV\_CNN\cite{9354642}} & \multirow{1}{*}{\centering SPSA,$L_\infty$,$\epsilon$=0.3} & \multirow{1}{*}{0.993} & \multirow{1}{*}{\centering 0.550} & \multirow{1}{*}{\centering 0.543} & \multirow{1}{*}{\centering 0.594} & \multirow{1}{*}{\centering 0.557} & \multirow{1}{*}{\centering 0.524} & \multirow{1}{*}{\centering 0.654} & \multirow{1}{*}{\centering 0.904} & \multirow{1}{*}{\centering \textbf{0.929}} \\
\midrule[0.5pt]
\multirow{1}{*}{\centering Lightweight\_FVCNN\cite{shen2021finger}} & \multirow{1}{*}{\centering SPSA,$L_\infty$,$\epsilon$=0.3} & \multirow{1}{*}{0.885} & \multirow{1}{*}{\centering 0.308} & \multirow{1}{*}{\centering 0.499} & \multirow{1}{*}{\centering 0.523} & \multirow{1}{*}{\centering 0.443} & \multirow{1}{*}{\centering 0.467} & \multirow{1}{*}{\centering 0.593 }& \multirow{1}{*}{\centering 0.837} & \multirow{1}{*}{\centering \textbf{0.851}} \\
\midrule[0.5pt]
\multirow{1}{*}{\centering FVRAS\_Net\cite{yang2020fvras}}  & \multirow{1}{*}{\centering SPSA,$L_\infty$,$\epsilon$=0.3} & \multirow{1}{*}{0.992} & \multirow{1}{*}{\centering 0.733} & \multirow{1}{*}{\centering 0.873} & \multirow{1}{*}{\centering 0.606} & \multirow{1}{*}{\centering 0.771} & \multirow{1}{*}{\centering 0.781} & \multirow{1}{*}{\centering 0.867} & \multirow{1}{*}{\centering 0.970} & \multirow{1}{*}{\centering \textbf{ {0.975}}} \\
\midrule[0.5pt]
\multirow{1}{*}{\centering DefenseGAN\_ModelB\cite{samangouei2018defense}} & \multirow{1}{*}{\centering SPSA,$L_\infty$,$\epsilon$=0.3}   & \multirow{1}{*}{0.914} & \multirow{1}{*}{\centering 0.449} & \multirow{1}{*}{\centering 0.757} & \multirow{1}{*}{\centering 0.549} & \multirow{1}{*}{\centering 0.535} & \multirow{1}{*}{\centering 0.503} & \multirow{1}{*}{\centering 0.719} & \multirow{1}{*}{\centering 0.872} & \multirow{1}{*}{\centering \textbf{0.886}} \\
\midrule[0.5pt]
\multirow{1}{*}{Res2Net50\cite{gao2019res2net}} & \multirow{1}{*}{\centering SPSA,$L_\infty$,$\epsilon$=0.3} & \multirow{1}{*}{0.993} & \multirow{1}{*}{\centering 0.295} & \multirow{1}{*}{\centering 0.457} & \multirow{1}{*}{\centering 0.563} & \multirow{1}{*}{\centering 0.399} & \multirow{1}{*}{\centering 0.336} & \multirow{1}{*}{\centering 0.665} & \multirow{1}{*}{\centering \textbf{0.970}} & \multirow{1}{*}{\centering 0.963} \\
\midrule[0.5pt]
\multirow{1}{*}{Vit\cite{dosovitskiy2010image}} & \multirow{1}{*}{\centering SPSA,$L_\infty$,$\epsilon$=0.3} & \multirow{1}{*}{0.900} & \multirow{1}{*}{\centering 0.558} & \multirow{1}{*}{\centering 0.730} & \multirow{1}{*}{\centering 0.590} & \multirow{1}{*}{\centering 0.580} & \multirow{1}{*}{\centering 0.539} & \multirow{1}{*}{\centering 0.677} & \multirow{1}{*}{\centering 0.872} & \multirow{1}{*}{\centering \textbf{0.892}} \\
\midrule[0.5pt]
\multirow{1}{*}{SwinTransformer\_v2\cite{liu2022swin}}    & \multirow{1}{*}{\centering SPSA,$L_\infty$,$\epsilon$=0.3} & \multirow{1}{*}{0.991} & \multirow{1}{*}{\centering 0.011} & \multirow{1}{*}{\centering 0.877} & \multirow{1}{*}{\centering 0.699} & \multirow{1}{*}{\centering 0.615} & \multirow{1}{*}{\centering 0.495} & \multirow{1}{*}{\centering 0.744} & \multirow{1}{*}{\centering 0.830} & \multirow{1}{*}{\centering \textbf{0.969}} \\
\midrule[0.5pt]
\multicolumn{2}{c}{Average / Gain} \vspace{5pt} \rule{0pt}{15pt} & 0.956 & 0.380 & 0.641 / 0.261 & 0.570 / 0.190 & 0.518 / {0.138} & 0.530 / 0.150 & 0.702 / 0.322 & 0.872 / 0.492 & \textbf{0.912} / \textbf{\textcolor{green}{0.532}} \\ 
\bottomrule[1pt]
\end{tabular}
}
\label{tab4}
\end{table*}
We observe that our MsMemoryGAN significantly outperforms existing defense methods in terms of recognition accuracy which reaches the highest accuracies, i.e. 0.981 and 0.982 on the two datasets, which are higher than those reached by MemAE \cite{2020Memorizing}, VeinGuard \cite{li2023transformer}, DefenseGAN \cite{samangouei2018defense}, Magnet \cite{meng2017magnet}, MemoryDefense\cite{adhikarla2022memory}, and DiffPure \cite{nie2022diffusion}, achieving thereby a new state-of-the-art. This performance may be attributed to the following facts:
\begin{itemize}
    \item The MsMemoryGAN's encoders extract feature representations at different scales from the images to reconstruct high-quality images. As shown in Fig. \ref{visual}, the local detail texture and global configurations of vein patterns are retained in the reconstructed images; 
    \item Our learnable metric enables the memory module to effectively find a combination of the most relevant normal patterns for a given input as its purifier version, which will be input into the decoder for reconstruction, thereby filtering perturbations. Also, the multi-scale framework allows the memory module to perform purification at different scales, which effectively removes the perturbations and reconstructs cleaned images;
    \item We combine feature perceptual losses, image reconstruction loss, and adversarial loss, for MsMemoryGAN training, to promote keeping semantic information in the reconstructed samples, resulting in high recognition accuracy.
\end{itemize}
By contrast, Magnet \cite{meng2017magnet} achieves promising results but shows poor performance in defending vein recognition models against attacks. This is due to the Magnet being an Autoencoder-based defense model. The Autoencoder aims to obtain a compressed encoding from the input and a decoder reconstructs the data from all the latent codes instead of fewer latent codes, which forces the network to extract the representative patterns of high-dimensional data. Therefore, the Autoencoder sometimes achieves good “generalization”  so that it can also reconstruct well the adversarial images, which is supported by \cite{2018Deep,2020Memorizing}.  
Similarly, VeinGuard is essentially an Autoencoder, so the perturbations may be reconstructed in the purified image, which degrades recognition accuracy. 
Defense-GAN relies on a traditional GAN which fails to learn an adequate distribution of normal data, which results in low-quality reconstructed images, with larger differences between the reconstructed and the original images (Fig. \ref{visual}(I)). 
The memory module in MemAE and MemoryDefense may solve this problem by using a few latent codes in the memory module for reconstruction. However, they only achieve image recognition at a single scale. Moreover, as the pixel loss instead of perceptual loss is used for model training, this results in blurred reconstructed images ((Fig. \ref{visual}(d)) and (Fig. \ref{visual}(f))).
To effectively filter the perturbations, the DiffPure model adds noise to adversarial examples by following the forward process with a small diffusion time step and recovering cleaned images by solving the reverse stochastic differential equation (SDE). It is very difficult, however, to determine the amount of noise to add during the forward process, so the reconstructed image
is still blurred, as shown in Fig. \ref{visual}(h).

\subsubsection{Black Box Attack Results}
In this section, we show the experimental results of different recognition models to assess the performance of our approach for black-box attacks, where the attacker cannot directly access the details of the target model and defense strategy, but can only observe and manipulate the model through its inputs and outputs. The attacker, therefore, will use the output of the model for a given input to generate adversarial samples. White-box attack methods, such as FGSM, can be used for black-box attacks, but they often require specially designed alternative models or even access to some of the training data. In recent years, the state-of-the-art SPSA approach \cite{uesato2018adversarial} was shown to perform an efficient black-box attack on the target model with only a finite number of model queries. In our experiments, we first apply SPSA to generate adversarial samples, which are then purified by the various defense methods. The cleaned images are then input to the recognition models, the recognition accuracy of which on the two datasets are reported in Table \ref{tab3} and Table \ref{tab4}, respectively.

From Table \ref{tab3} and Table \ref{tab4}, we observe that our MsMemoryGAN still shows the highest performance under black-box attacks, which is consistent with the results in Table \ref{tab1} and Table \ref{tab2} under white-box attacks. After filtering the perturbations by MsMemoryGAN,  the various recognition models obtain a significant accuracy improvement, {about 26.5\% average  improvement on dataset TJU\_PV, and about 53.2\% improvement on dataset PolyU M\_N.}

\section{Conclusion}
In this paper, we have proposed MsMemoryGAN, a novel defense model against palm-vein adversarial attacks. First, we proposed a multi-scale AutoEncoder to improve feature representation capacity. Then, we designed a memory module with a learnable metric for memory addressing. 
Finally, we proposed a perceptional loss and an adversarial loss for training. 
As our MsMemoryGAN reconstructs an adversarial sample by retrieving its typical normal patterns from memory, it effectively removes adversarial perturbations. Our experiments on two public vein datasets with different attacks demonstrate that our MsMemoryGAN is the best at defending against adversarial attacks and achieves a new state-of-the-art.

\bibliographystyle{unsrt}
\bibliography{bibtex}
\vspace{-15 mm}

\end{document}